\documentclass{sigchi-ext}
\usepackage[T1]{fontenc}
\usepackage{textcomp}
\usepackage[scaled=.92]{helvet} 
\usepackage{graphicx} 
\usepackage{balance}  
\usepackage{booktabs} 
\usepackage{ccicons}  
\usepackage{multirow}
\usepackage{gensymb}
\usepackage{soul}
\sethlcolor{red}
\renewcommand{\hl}[1]{#1}  
\newcommand{\figureraisebox}[1]{\raisebox{0.3cm}{#1}}

\def\plainkeywords{Mobile Robot; Video Summary; Senior; Indoor Activity}

\title{A Mobile Robot Generating Video Summaries of Seniors' Indoor Activities}

\numberofauthors{4}
\author{%
  \alignauthor{%
    \textbf{Chih-Yuan Yang}\\
    \affaddr{National Taiwan University} \\
    \affaddr{Taipei, Taiwan} \\
    \email{yangchihyuan@csie.ntu.edu.tw} }
    \alignauthor{%
    \textbf{Srenavis Varadaraj}\\	
    \affaddr{Intel Labs, Intel Technologies India Pvt. Ltd.}\\
    \affaddr{Bangalore, India}\\
    \email{srenivas.varadarajan@intel.com} } \vfil
    \alignauthor{%
    \textbf{Heeseung Yun}\\			
    \affaddr{Seoul National University}\\
    \affaddr{Seoul, Korea}\\
    \email{terry9772@snu.ac.kr} }
    \alignauthor{%
    \textbf{Jane Yung-jen Hsu}\\
    \affaddr{National Taiwan University} \\
    \affaddr{Taipei, Taiwan} \\
    \email{yjhsu@csie.ntu.edu.tw} }
     }

\begin{document}


\maketitle

\RaggedRight{} 

\begin{abstract}
We develop a system which generates summaries from seniors' indoor-activity videos captured by a social robot to help remote family members know their seniors' daily activities at home. Unlike the traditional video summarization datasets, indoor videos captured from a moving robot poses additional challenges, namely, (i) the video sequences are very long (ii) a significant number of video-frames contain no-subject or with subjects at ill-posed locations and scales (iii) most of the well-posed frames contain highly redundant information. To address this problem, we propose to \hl{exploit} pose estimation \hl{for detecting} people in frames\hl{. This guides the robot} to follow the user and capture effective videos. We use person identification to distinguish a target senior from other people. We \hl{also make use of} action recognition to analyze seniors' major activities at different moments, and develop a video summarization method to select diverse and representative keyframes as summaries.  
\end{abstract}



\section{Introduction}
\justify
With a large portion of the population becoming aged, we aim to investigate the feasibility of applying video summarization techniques using a social robot to help family members care about seniors living alone. Numerous video summarization methods and interactive robots have been independently developed, but they were never put together. Thus, we investigate their limitations and propose our solution to the widely growing demand.

\begin{marginfigure}[0pc]
  \begin{minipage}{\marginparwidth}
    \centering
    \includegraphics[width=0.9\marginparwidth]{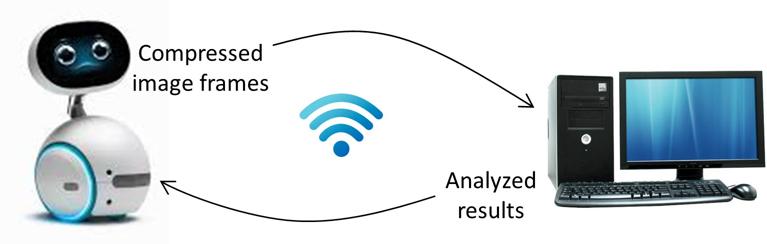}
    \caption{System architecture. Due to the limited computational resources available on a robot, we extend them by a high-performance computer. We transmit data between the robot and the computer via the wireless connection to ensure the robot's moving capability.}
    ~\label{fig:HardwareDevices}
  \end{minipage}
\end{marginfigure}
\begin{marginfigure}[0pc]
  \begin{minipage}{\marginparwidth}
    \centering
	\begin{tabular}{@{}c@{\hspace{0.02\columnwidth}}c@{\hspace{0.02\columnwidth}}c@{}}
		\includegraphics[width=0.32\columnwidth]{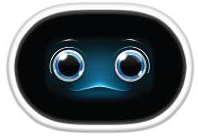} &
		\includegraphics[width=0.32\columnwidth]{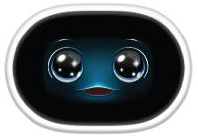} &
		\includegraphics[width=0.32\columnwidth]{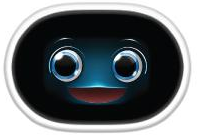} \\
		(a) & (b) & (c) \\
		\includegraphics[width=0.32\columnwidth]{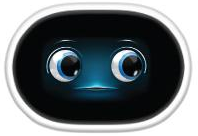} &
		\includegraphics[width=0.32\columnwidth]{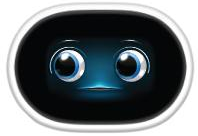} & \\
		 (d) & (e) &\\
	\end{tabular}
	\vspace{-0.2cm}
	\caption{Facial expressions shown on our robot. (a) default\_still, if no human is observed. (b) expecting, if the robot is seeing a user's back. (c) active, if the robot is seeing a user's eyes. (d)(e) aware\_left and aware\_right, when the robot is turning left or right to look for a user.}
	\label{fig:RobotFacialExpressions}
	\end{minipage}
\end{marginfigure}
\section{Related Work}
Our study covers the topics of mobile robots and video summarization because we use a mobile robot to capture videos and analyze the videos to generate summaries.

Autonomous Mobile Robots are capable of navigating an uncontrolled environment and moving around to carry out given tasks. 
If those robots interact and communicate with humans and their tasks are to improve the quality of users' life, they belong to social robots. Many studies have shown that social robots are good tools to meet the elderly individual needs and requirements~\cite{Klamer10}. 
With the advance of technology, social robots have been extended from zoomorphic robots to humanoid mobile robots equipped with multiple sensors and advanced intelligence~\cite{Chen18_FG,Tan18_IJSR}.
In this paper, we assign a new task \hl{of generating video summaries to social robots} and develop a solution.

Video Summarization methods analyze videos to create summaries. It is part of machine learning and data mining. The main idea of summarization is to find a diverse and representative subset of the entire input data. Several summary formats are available including keyframes, skimmed videos, time-lapsed videos, and video synopsis. Among them, the keyframes are widely used because they are simple and ease to consume~\cite{Avila11,Zhang16_ECCV}.

\section{System Architecture}
We use a commercial robot Zenbo, as shown in Figure~\ref{fig:HardwareDevices}, to capture videos. It uses an ultra-low-voltage CPU to reduce power consumption, but the \hl{processing power of the CPU} is insufficient to analyze videos in real time. Thus, we transmit captured frames via Wi-Fi to a powerful computer to analyze them and return results to the robot.

The robot has a touch screen \hl{on} its head, which serves as an input UI and also shows facial expressions for human-computer interaction. The robot's camera is at the upper center boundary of the screen. The robot's OS is a modified Android system with a set of additional APIs to retrieve sensor data and control the robot's movements, neck gestures, and facial expressions. There are 24 built-in expressions rendered by OpenGL. We select five of them, as shown in Figure~\ref{fig:RobotFacialExpressions}, to interact with users.
Our software implementation includes three parts, an Android-based app running on the robot to transmit captured frames to a server and receive analyzed results from the server to control the robot's actions, a C++ program running on the server to analyze images, and an offline Python program to generate summaries by selecting keyframes from all captured frames.

\section{Human Detection}
\begin{figure}
	\centering
	\includegraphics[width=1\columnwidth]{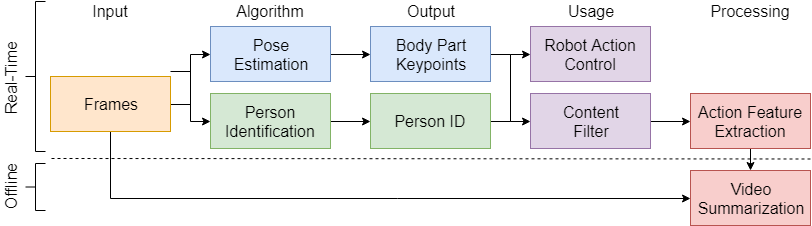}
	\vspace{-0.3cm}
	\caption{Server-side flowchart of data analysis}
	\label{fig:Flowchart}
	\vspace{-5pt}
\end{figure}

Because our summaries focus on human activities, we utilize existing human detection algorithms to find our human targets. Figure~\ref{fig:Flowchart} shows the relationship \hl{among} those algorithms. Given an input video frame, we use a real-time pose estimation algorithm OpenPose~\cite{Cao17_CVPR_OpenPose} to find people. We use a pose estimation method rather than a pedestrian or object detection method because we need human body landmark coordinates to control our robot's camera view angle. We want our robot to capture well-posed frames so we can generate high-quality summaries. In contrast, bounding boxes reported by a pedestrian or object detection method do not \hl{analyze} human body parts.
Because pose estimation methods do not \hl{discern} people's identities, we use a person identification method PSE-ECN~\cite{Safriaz18} to prevent false-positives and determine our target person among people on an image.
\begin{marginfigure}[-7pc]
  \begin{minipage}{\marginparwidth}
    \centering
    \includegraphics[width=0.5\marginparwidth]{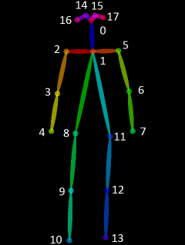}
    \caption{Landmark points.}~\label{fig:Landmarks}
  \end{minipage}
\end{marginfigure}
\begin{marginfigure}[0pc]
  \begin{minipage}{\marginparwidth}
	\centering
	\includegraphics[width=1\columnwidth]{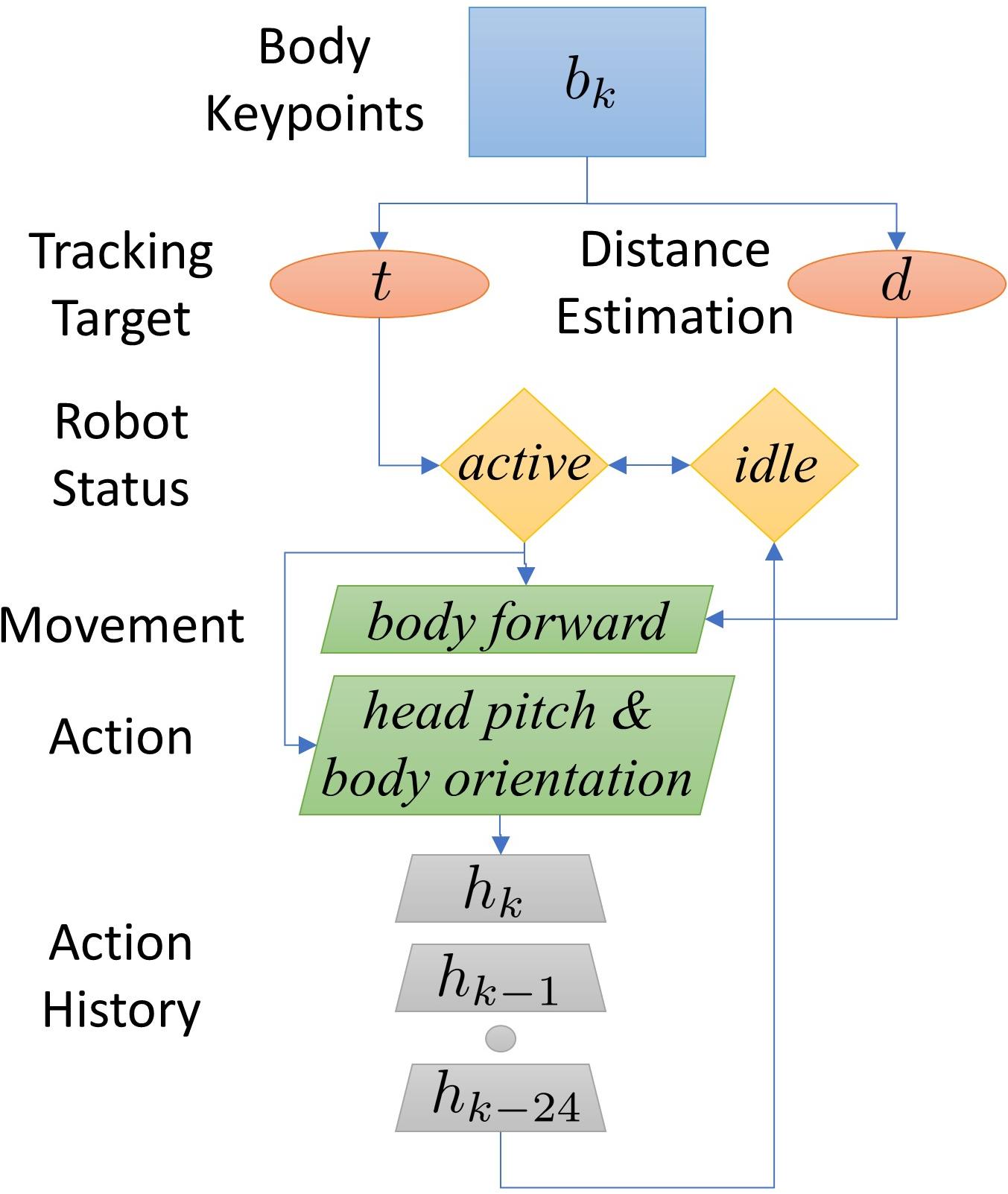}
	\caption{Robot-side data flow for movement control.}
	\label{fig:MovingMode}
  \end{minipage}
\end{marginfigure}
\section{Rule-based Robot Movement Control}
OpenPose reports 18 body landmark coordinates as shown in Figure~\ref{fig:Landmarks}, but some may be unavailable if the body parts are invisible. Regarding our target person's coordinates, if any facial landmark point is available, we adjust the robot's body orientation and neck pitch angle to change the robot's camera view angle and move the landmark point(s) towards the central upper part of the upcoming frames to create well-posed ones for summarization. If all facial landmark points are invisible but the neck point is visible, we raise the robot's head to look for a face. In addition, we also use landmark points to estimate the distance between our robot and the target user to control the \hl{robot's} forward movement. We compute the average distances from the neck point to the two hip points because they are the longest connected distances and they distort less than other body parts do in different poses. We move our robot toward the target person until 2 meters apart, which is a proper distance to take images with adequate human size.
When our robot cannot \hl{detect} a person, it will turn left or right 30 degrees depending on the last position of a person visible in past frames. If the robot turns around but still cannot \hl{detect} a person, we set the robot's neck pitch angle 15 degree vertically because this degree ensures the robot \hl{to detect} distant people. If the robot turns around twice but still does not \hl{detect} a person, we let the robot wait in an idle mode for 15 minutes to save energy until a person appears in front of its camera. Figure~\ref{fig:MovingMode} shows the overall flow of movement control.

\section{Content Filter}
\begin{marginfigure}[0pc]
  \begin{minipage}{\marginparwidth}
    \centering
	\begin{tabular}{c}
	\includegraphics[width=0.4\marginparwidth]{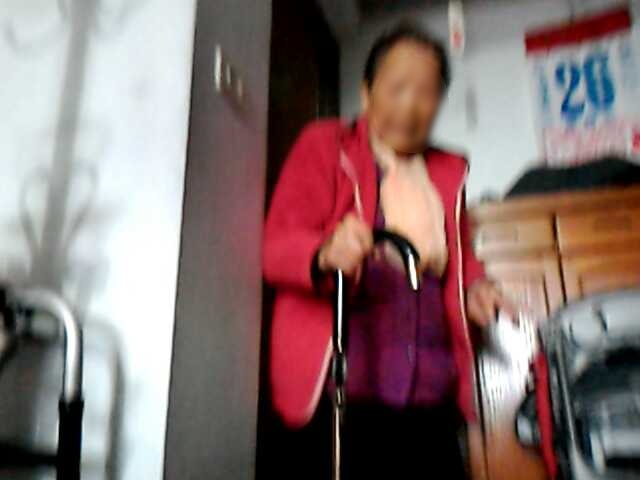} \\
	(a) Blurred \\
	\includegraphics[width=0.4\marginparwidth]{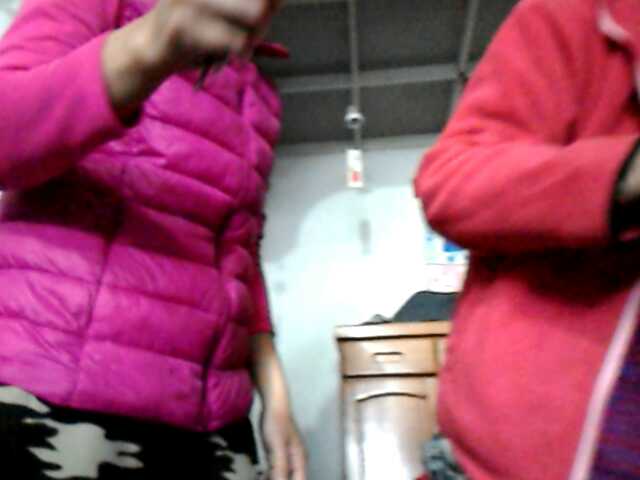} \\
	(b) Eyes invisible \\
	\includegraphics[width=0.4\marginparwidth]{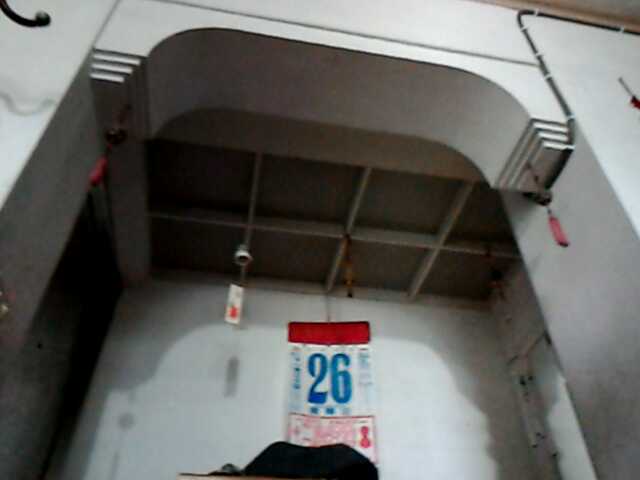} \\
	(c) People absent \\
	\includegraphics[width=0.4\marginparwidth]{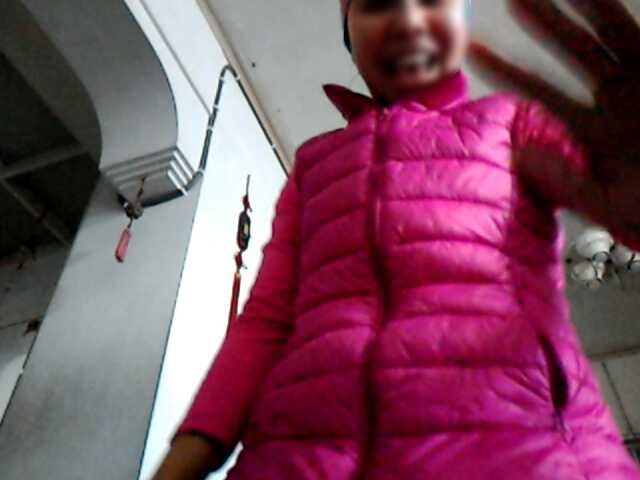} \\
	(d) Forehead cropped\\
	\includegraphics[width=0.4\marginparwidth]{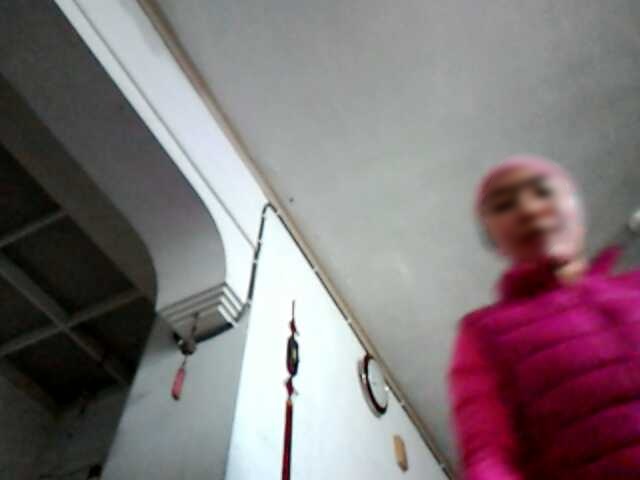} \\
	(e) People at corners \\
	\includegraphics[width=0.4\marginparwidth]{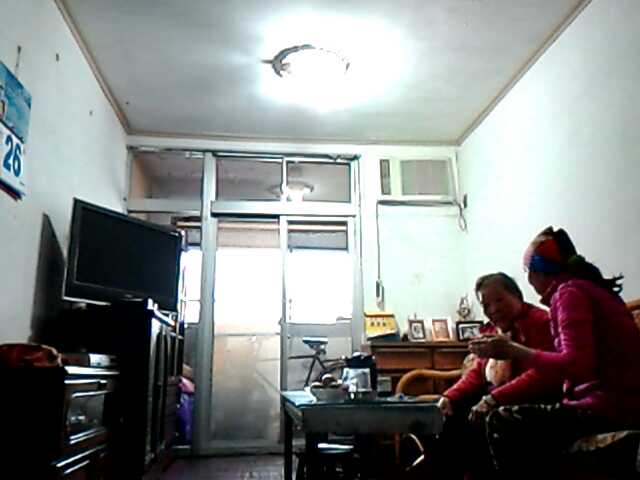} \\
	(f) People too small\\
	\end{tabular}
    \caption{Examples of ill-posed frames, which are removed by our content filter.}~\label{fig:IllPosedFrames}
  \end{minipage}
\end{marginfigure}

Our images are captured from a moving camera in an indoor space so many of them are blurred and improper to be selected as summaries. To remove them, we use the variance-of-Laplacian~\cite{PechPacheco2000_DiatomAI} method and set a threshold. We also use the aforementioned OpenPose and PSE-ECN methods to ignore ill-posed frames, including the ones without people or with people but too small, cropped, at corners, or whose faces are invisible. Figure~\ref{fig:IllPosedFrames} show\hl{s} examples for 6 cases.

\section{Video Summarization}
We use keyframes as our summary format due to its efficiency to consume information at a glance. We propose a method to select keyframes by temporally clustering well-posed frames and then selecting one representative frame out of a cluster in terms of human actions because we expect the summary to not only diversely cover seniors' daily activities but also show the representative ones. 
To do it, let $\{v_i\}$, $i \in$ (1,$n$), be the well-posed frames and $\{t_i\}$ be their timestamps, $t_i < t_{i+1}$.
Let $k$ be the number of keyframes in our summary. We group $\{v_i\}$ into at least $k$ clusters $\{C_j\}$, $j \in$ (1,$m$) and $m \geq k$. 
Initially we let $C_1$ contain the first frame $v_1$. For any other frames $v_i$, $i \in$ (2,$n$), we assign their clusters by the temporal difference from its previous frame, i.e.,
assume $v_{i-1} \in C_j$, we assign
$
v_i \in \left\{ 
\begin{array}{ll} 
C_j & \mbox{if $t_{i} - t_{i-1} < h$}; \\
C_{j+1} & \mbox{else},
\end{array} 
\right.
$
where $h$ is a temporal gap threshold.
In order to produce at least $k$ clusters, We iteratively adjust
$
h = \left\{ 
\begin{array}{ll} 
2h & \mbox{if $m \ge k$}; \\
\frac{h}{2} & \mbox{else,}
\end{array} 
\right.
$
until $m(h) \ge k$ and $m(2h) < k$ where $m(h)$ is the number of clusters determined by $h$. If $m(h) > k$, we disregard the small $m(h)-k$ clusters and only use the $k$ large ones.

We extract frame features as the probabilities of 157 predefined indoor actions generated by a SqueezeNet~\cite{Iandola16_SqueezeNet} model pre-trained on the Charades dataset~\cite{Sigurdsson16}, which aims to recognize human actions from a single frame. We compute the mean features of a cluster, and select the frame with the closest distance to the cluster mean, as the \hl{cluster's keyframe}.

\begin{table}[t]
	\centering
	\begin{tabular}{l@{\hspace{2mm}}l@{\hspace{2mm}}l@{\hspace{2mm}}l}
		\hline
		Video & Subjects &  Duration & \#Frame (Total/well-posed)\\
		\hline
		\multirow{3}{*}{1} & male (79) & \multirow{3}{*}{6h 46m} & \multirow{3}{*}{198689 / 8093}\\
		& female (74) &  & \\
		& female (41) & & \\
		\hline
		\multirow{2}{*}{2} & female (94) & \multirow{2}{*}{1h 44m} & \multirow{2}{*}{50634 / 19971}\\
		&	female (31) & & \\
		\hline
		3 & female (70) & 7h 42m & 191183 / 11563 \\
		\hline
	\end{tabular}
	\caption{Statistics of experimental videos. The subjects' ages are shown beside their genders.}
	\label{table:ExperimentList}
	\vspace{-5pt}
\end{table}
\section{Experiments}
We conduct experiments in three families, and their statics are shown in Table~\ref{table:ExperimentList}. We set the keyframe number $k$ as 8 and the initial threshold $h$ as 60 seconds. We compare the proposed video summarization method with three existing ones: VSUMM~\cite{Avila11}, DPP~\cite{Gong14}, and DR-DSN~\cite{Zhou18_AAAI}. All of them are programmed in Python and run on a machine equipped with a 3.4GHz quadratic core CPU and their execution time is shown in Table~\ref{table:execution_time}. We use the publicly available code of OpenPose implemented in OpenVINO~\cite{OpenVINO} but temporally disable PSE-ECN because we have not fully integrated it into our program.
\begin{table}[t]
	\centering
	\begin{tabular}{l@{\hspace{6pt}}c@{\hspace{6pt}}c@{\hspace{6pt}}c@{\hspace{6pt}}c}
	\hline
	Method & VSUMM & DPP & DR-DSN & Proposed \\
	\hline
	Time & 11.05s & 11m 32s & 11m 27s & \textbf{0.18s} \\
	\hline
	\end{tabular}
	\caption{Execution time for video 2. The execution time for videos 1 and 3 is proportional to their frame numbers.}
	\label{table:execution_time}	
\end{table}

\begin{figure*}[t]
	\centering
	\begin{tabular}{@{}c@{\hspace{0.03\columnwidth}}r@{\hspace{0.005\columnwidth}}c@{\hspace{0.01\columnwidth}}c@{\hspace{0.01\columnwidth}}c@{\hspace{0.01\columnwidth}}c@{\hspace{0.01\columnwidth}}c@{\hspace{0.01\columnwidth}}c@{\hspace{0.01\columnwidth}}c@{\hspace{0.01\columnwidth}}c@{\hspace{0.01\columnwidth}}l}
		\multirow{5}{*}[0.05cm]{\includegraphics[width=0.6\columnwidth]{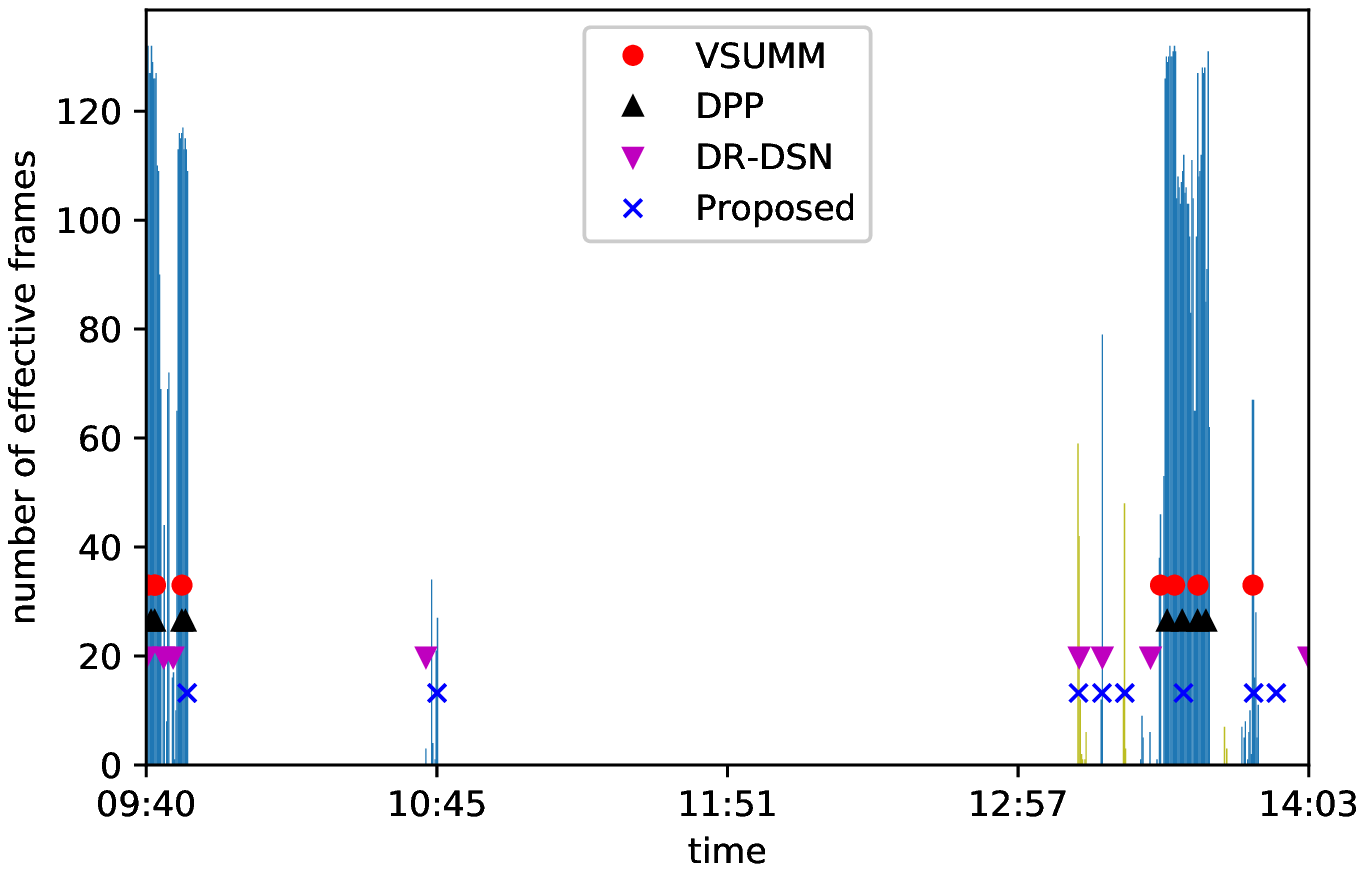}} &
		\figureraisebox{1} &
		\includegraphics[width=0.14\columnwidth]{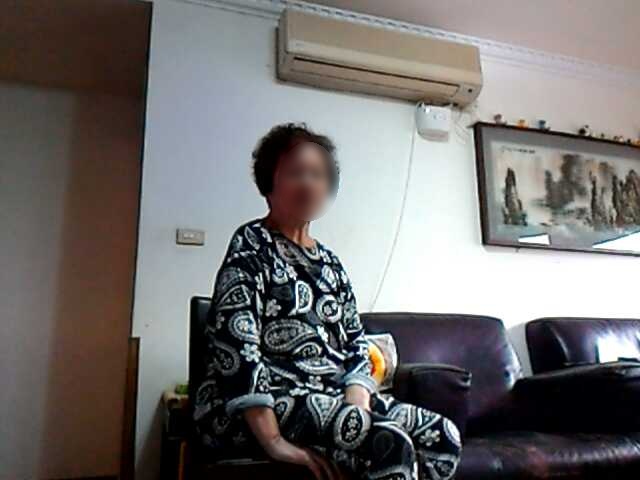} &
		\includegraphics[width=0.14\columnwidth]{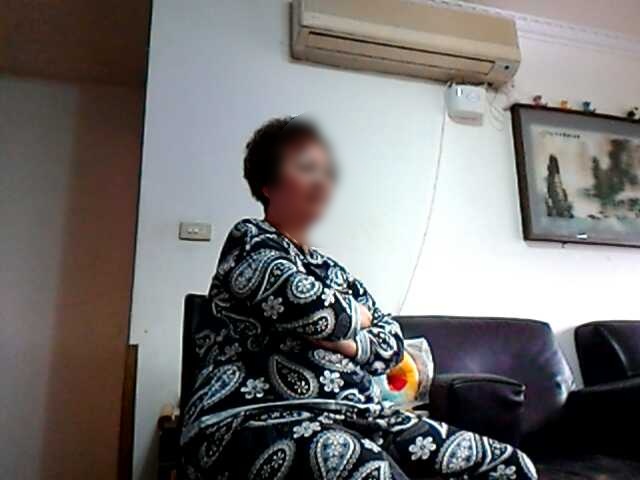} &
		\includegraphics[width=0.14\columnwidth]{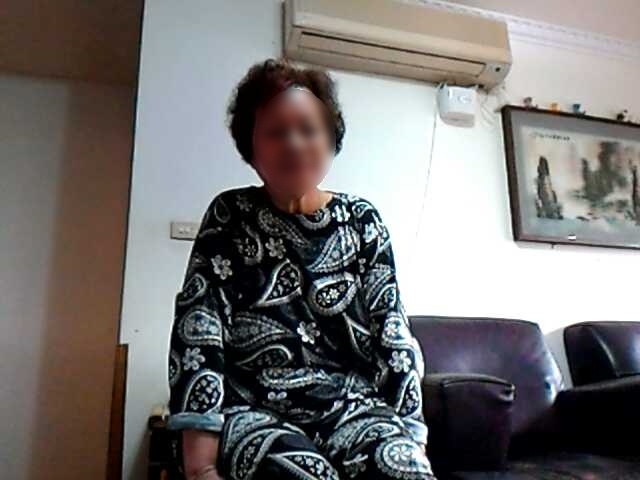} &
		\includegraphics[width=0.14\columnwidth]{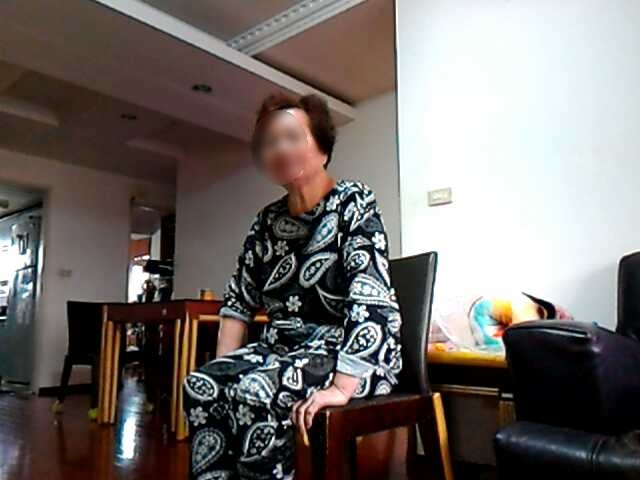} &
		\includegraphics[width=0.14\columnwidth]{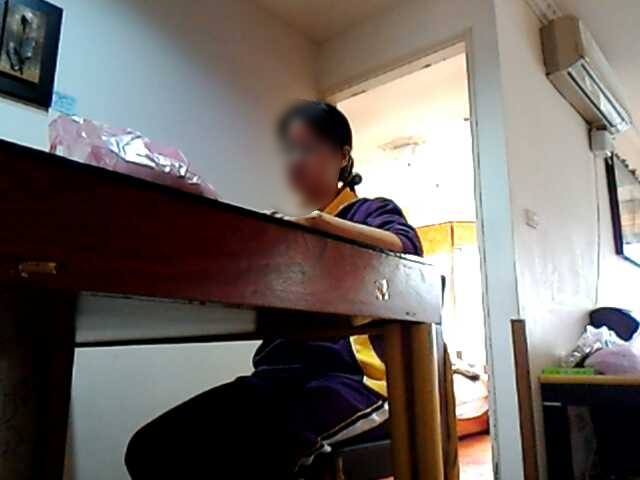} &
		\includegraphics[width=0.14\columnwidth]{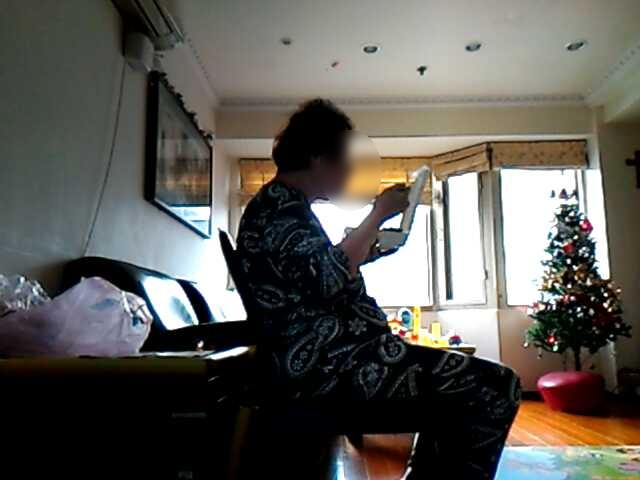} &
		\includegraphics[width=0.14\columnwidth]{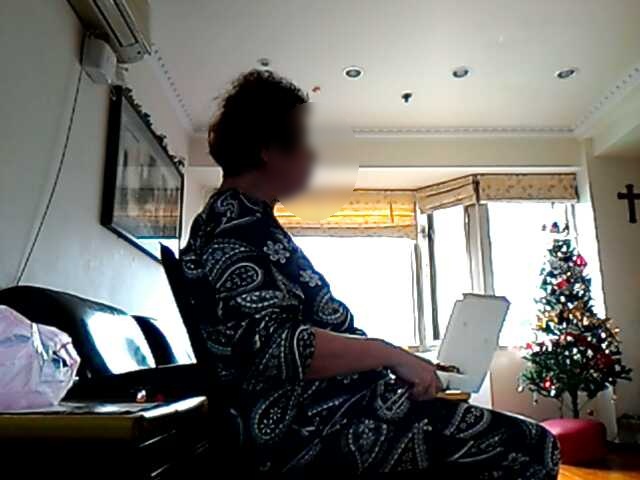} &
		\includegraphics[width=0.14\columnwidth]{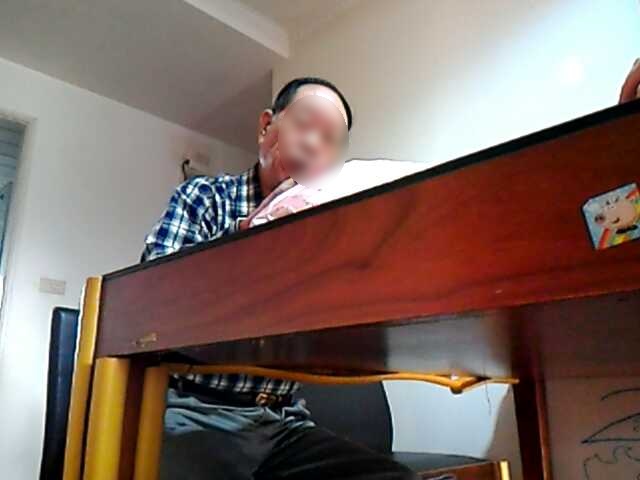} &
		\figureraisebox{VSUMM}\\
		&
		\figureraisebox{2} &
		\includegraphics[width=0.14\columnwidth]{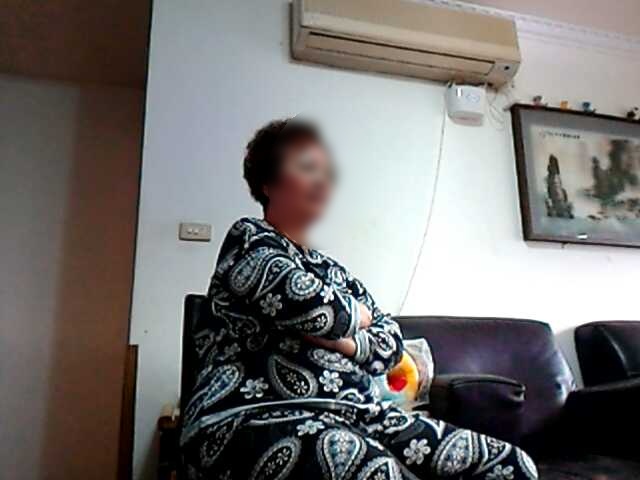} &
		\includegraphics[width=0.14\columnwidth]{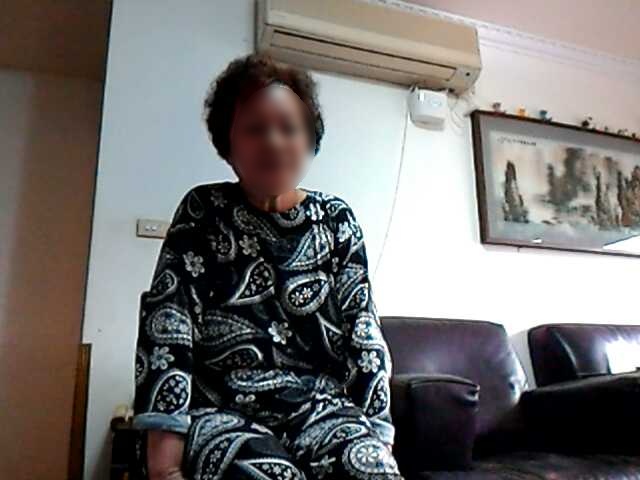} &
		\includegraphics[width=0.14\columnwidth]{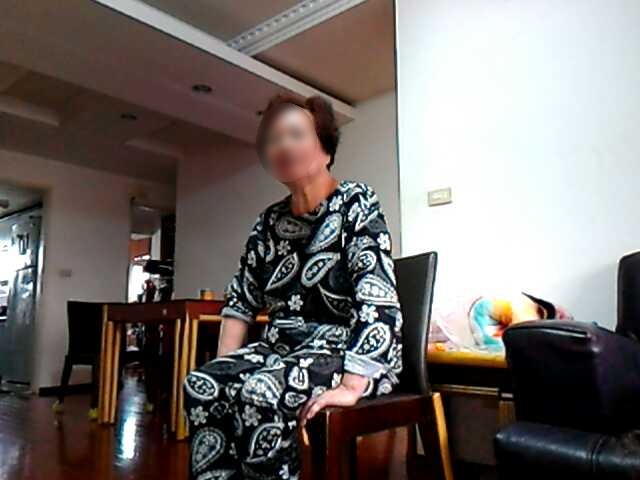} &
		\includegraphics[width=0.14\columnwidth]{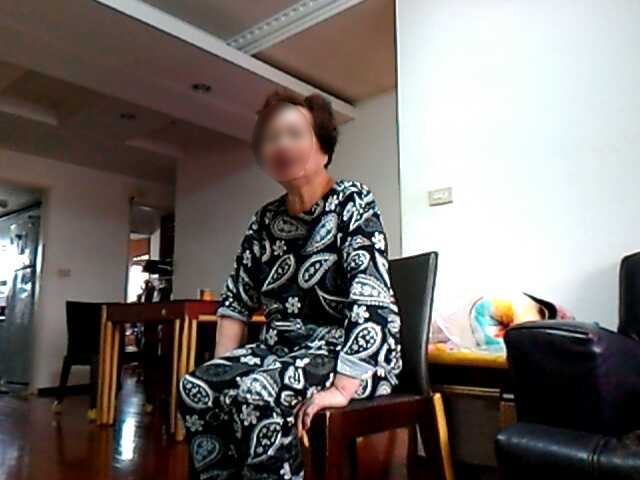} &
		\includegraphics[width=0.14\columnwidth]{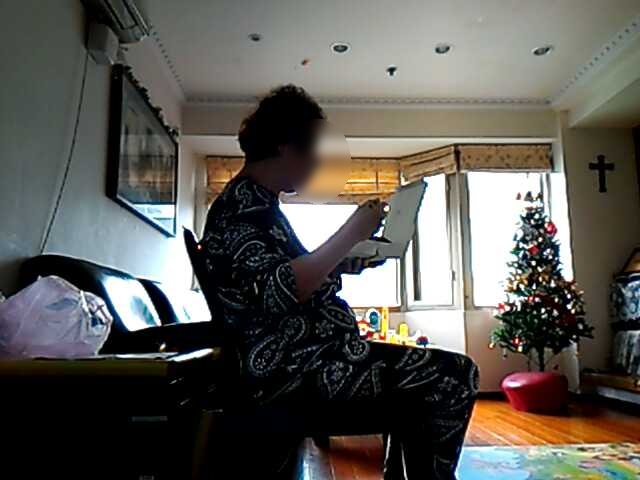} &
		\includegraphics[width=0.14\columnwidth]{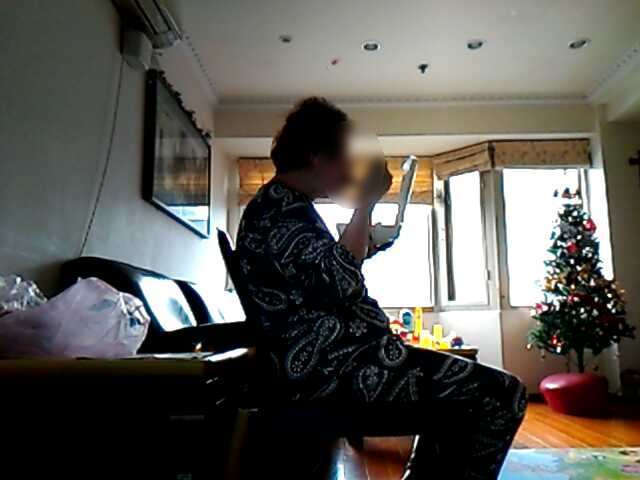} &
		\includegraphics[width=0.14\columnwidth]{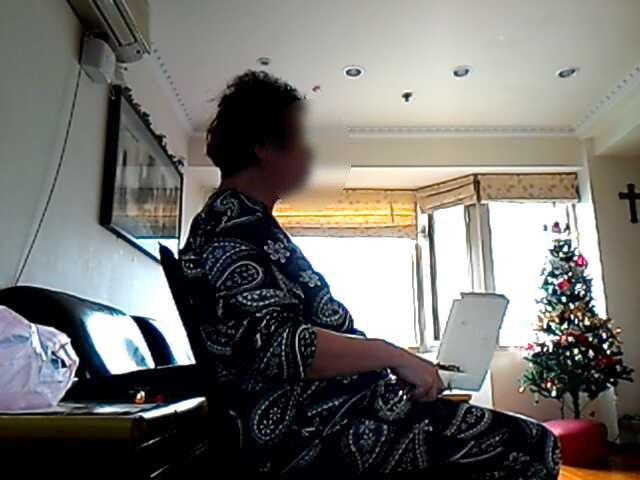} &
		\includegraphics[width=0.14\columnwidth]{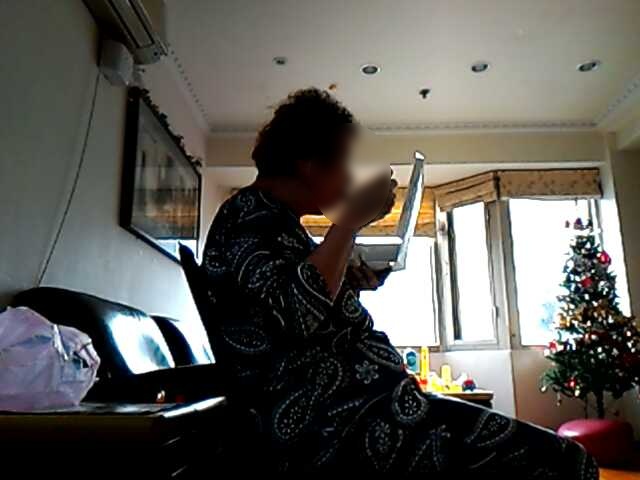} &
		\figureraisebox{DPP}\\
		&
		\figureraisebox{3} &
		\includegraphics[width=0.14\columnwidth]{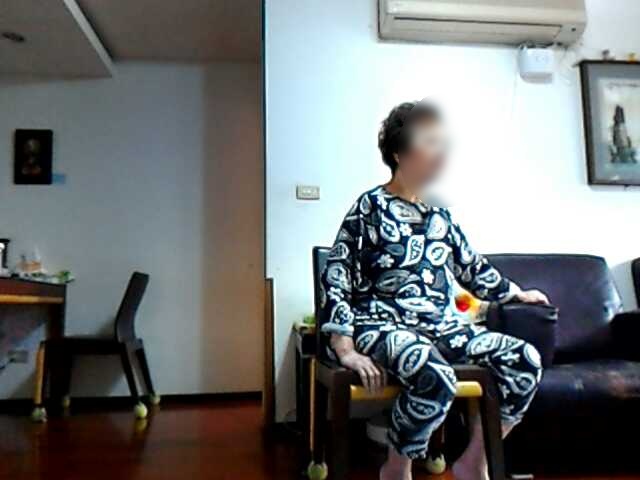} &
		\includegraphics[width=0.14\columnwidth]{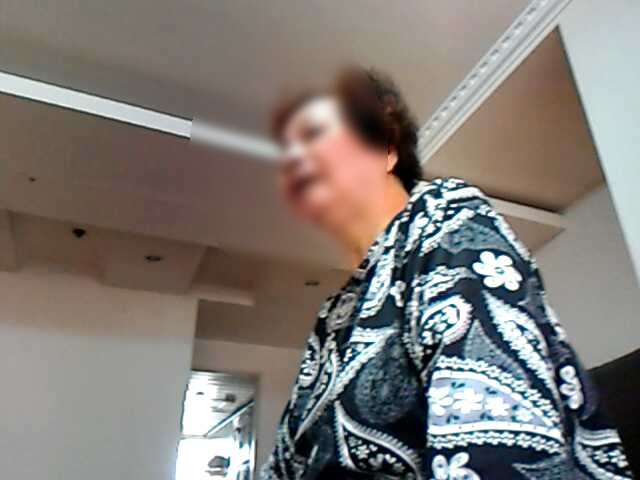} &
		\includegraphics[width=0.14\columnwidth]{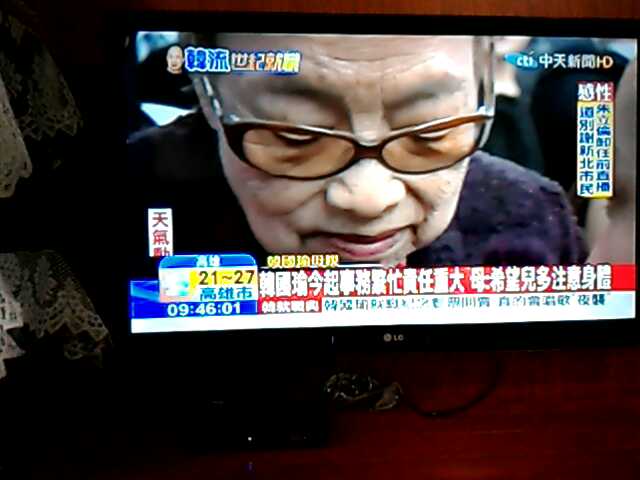} &
		\includegraphics[width=0.14\columnwidth]{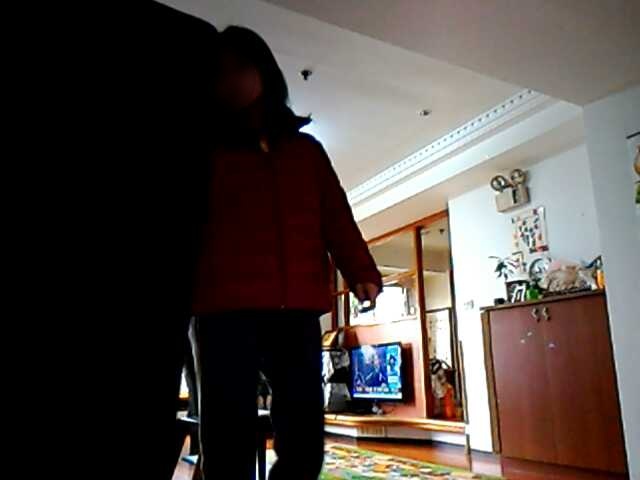} &
		\includegraphics[width=0.14\columnwidth]{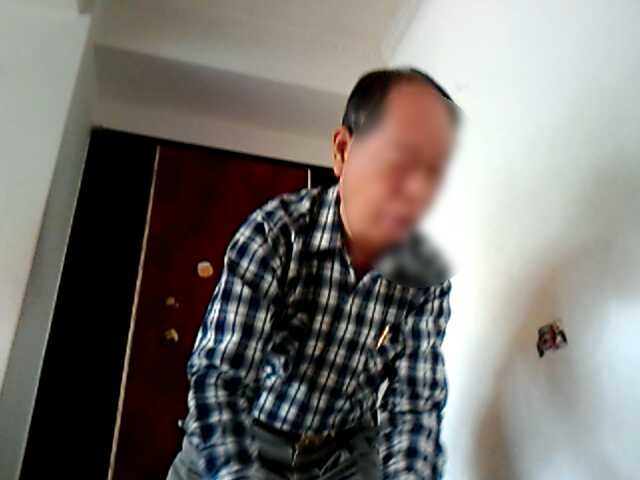} &
		\includegraphics[width=0.14\columnwidth]{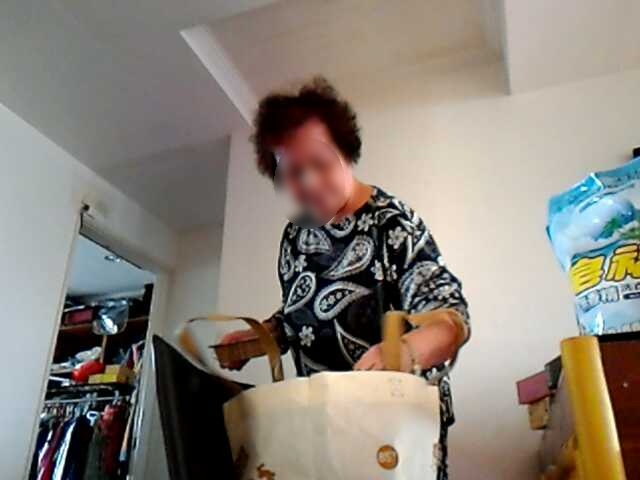} &
		\includegraphics[width=0.14\columnwidth]{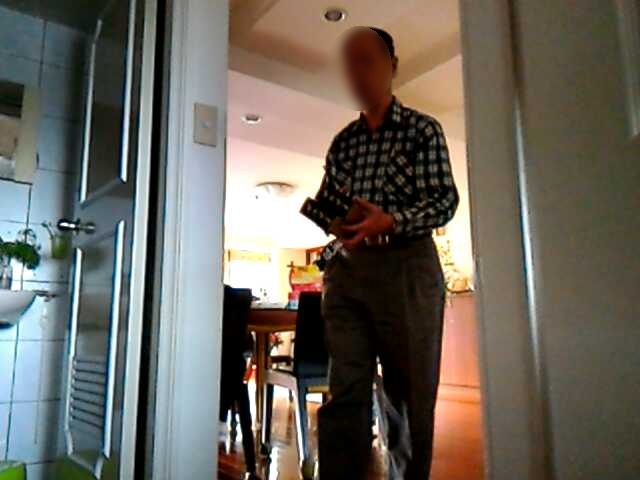} &
		\includegraphics[width=0.14\columnwidth]{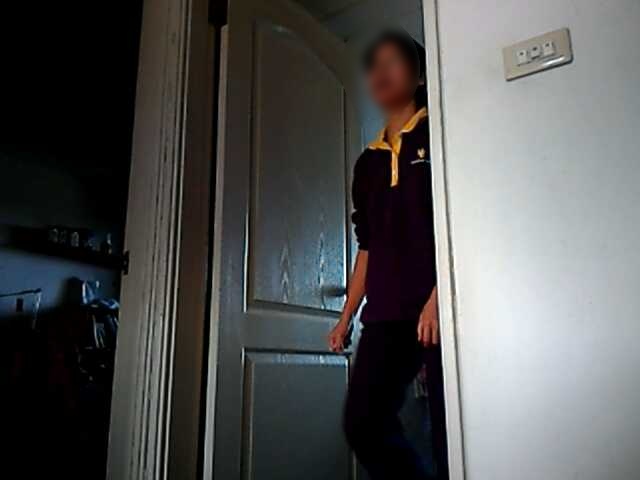} &
		\figureraisebox{DR-DSN}\\
		&
		\figureraisebox{4} &
		\includegraphics[width=0.14\columnwidth]{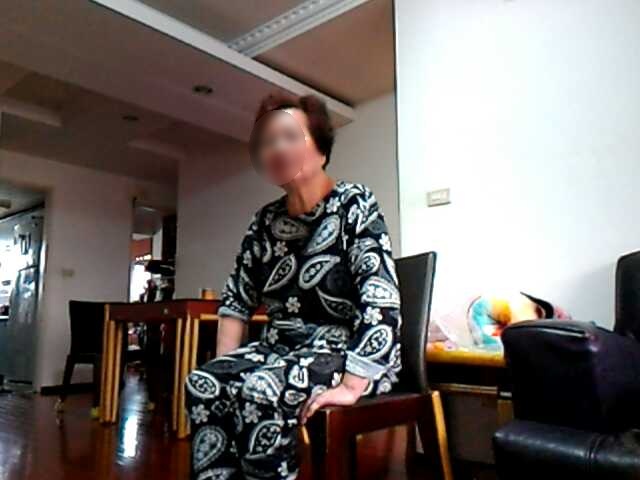} &
		\includegraphics[width=0.14\columnwidth]{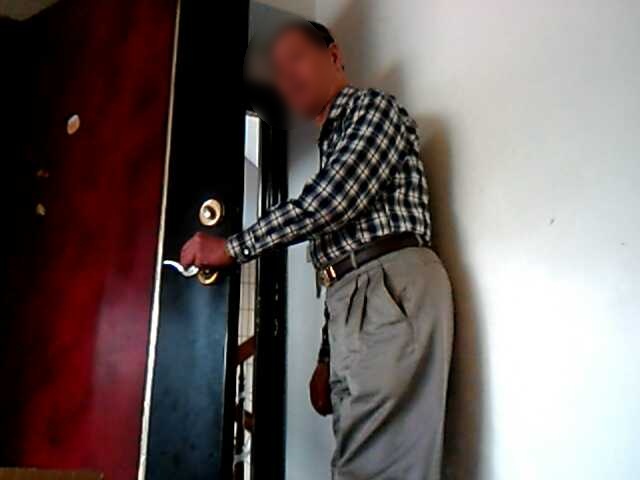} &
		\includegraphics[width=0.14\columnwidth]{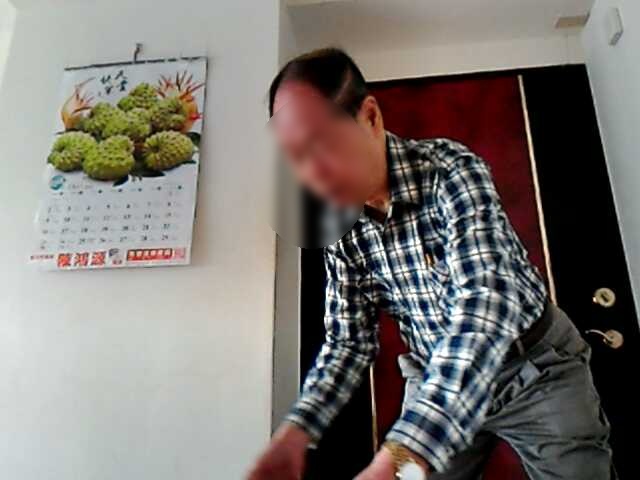} &
		\includegraphics[width=0.14\columnwidth]{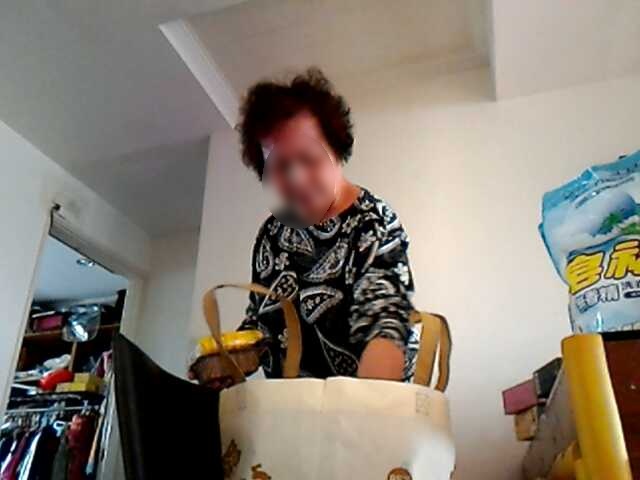} &
		\includegraphics[width=0.14\columnwidth]{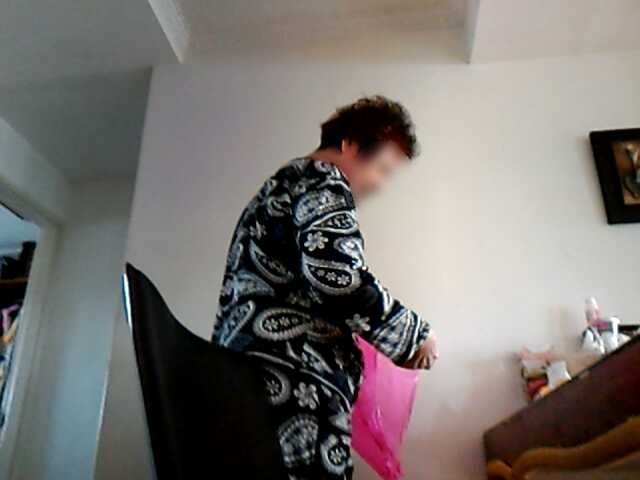} &
		\includegraphics[width=0.14\columnwidth]{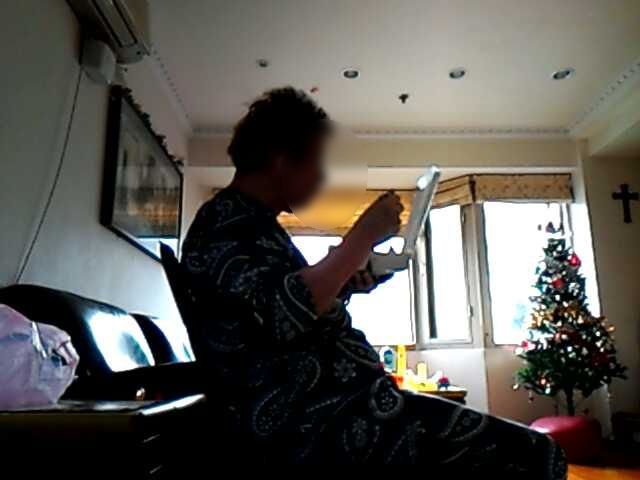} &
		\includegraphics[width=0.14\columnwidth]{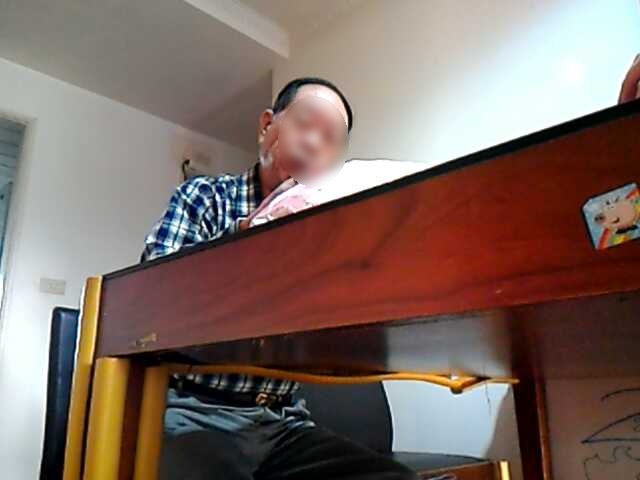} &
		\includegraphics[width=0.14\columnwidth]{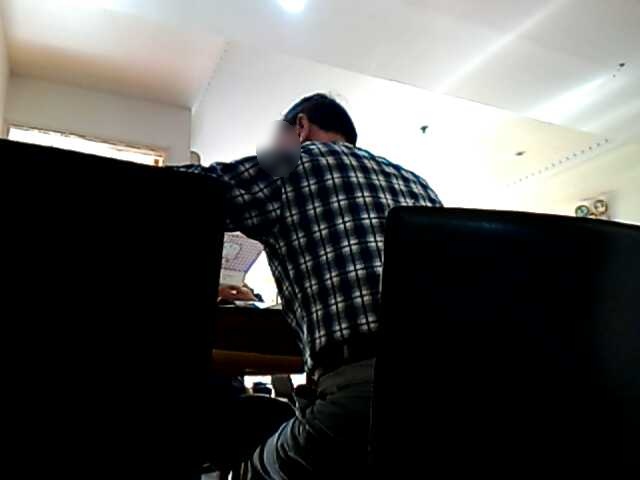} &
		\figureraisebox{Proposed}\\
		\rule{0pt}{0.8cm}\multirow{5}{*}[0.05cm]{\includegraphics[width=0.6\columnwidth]{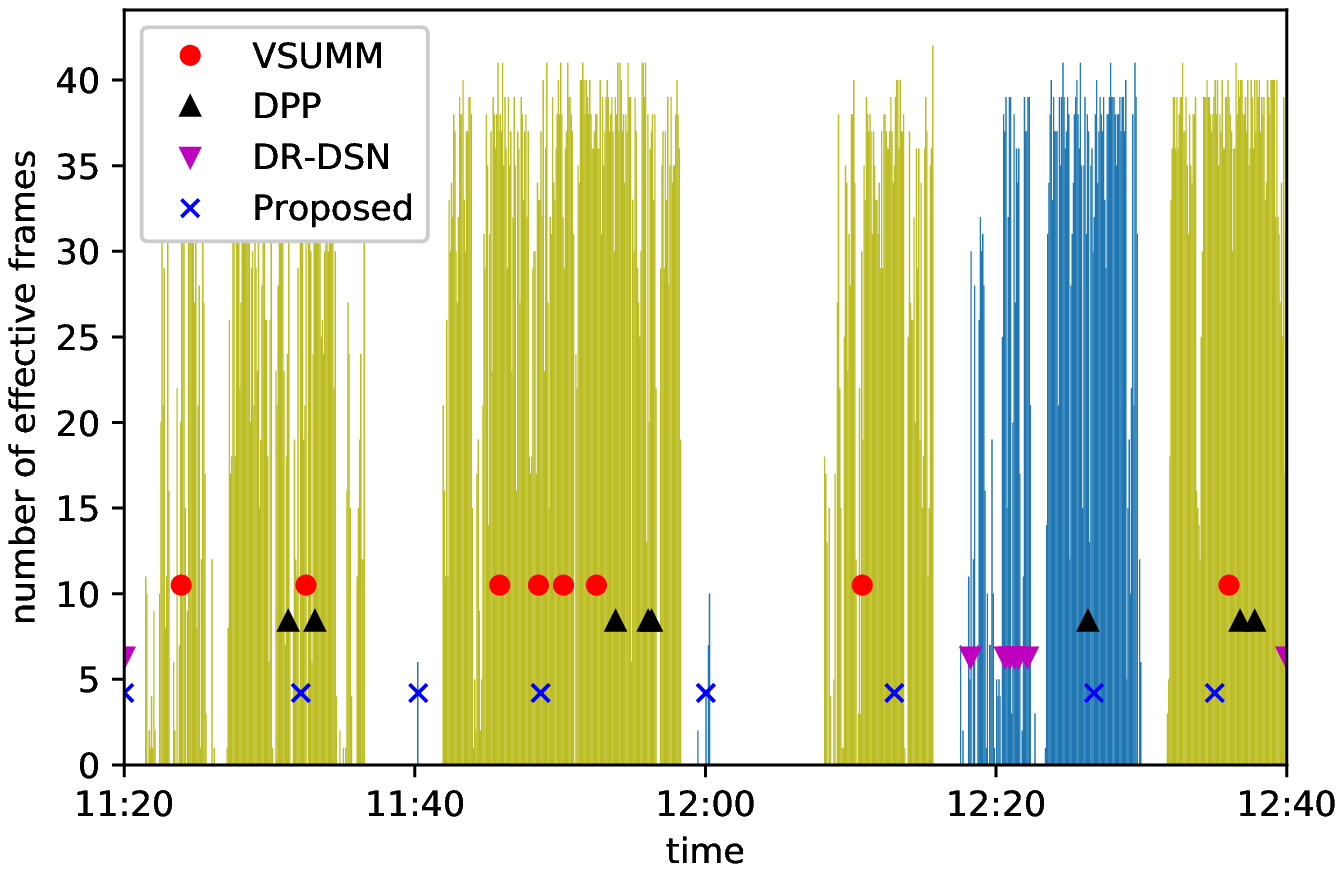}} &
		\figureraisebox{5} &
		\includegraphics[width=0.14\columnwidth]{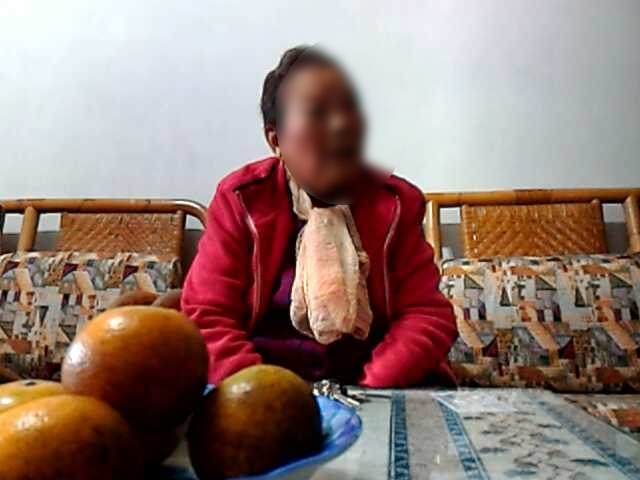} &
		\includegraphics[width=0.14\columnwidth]{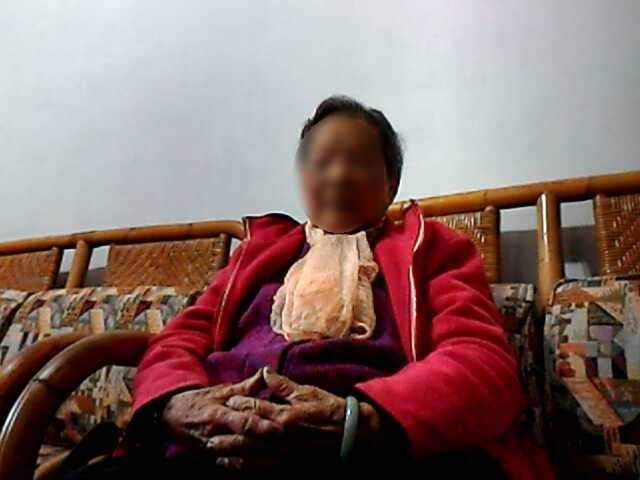} &
		\includegraphics[width=0.14\columnwidth]{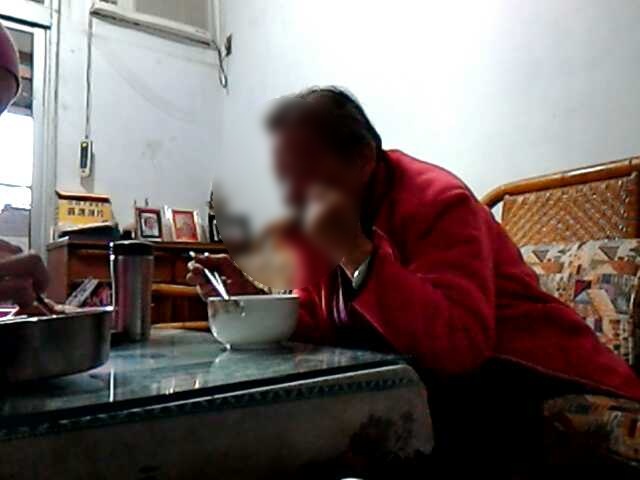} &
		\includegraphics[width=0.14\columnwidth]{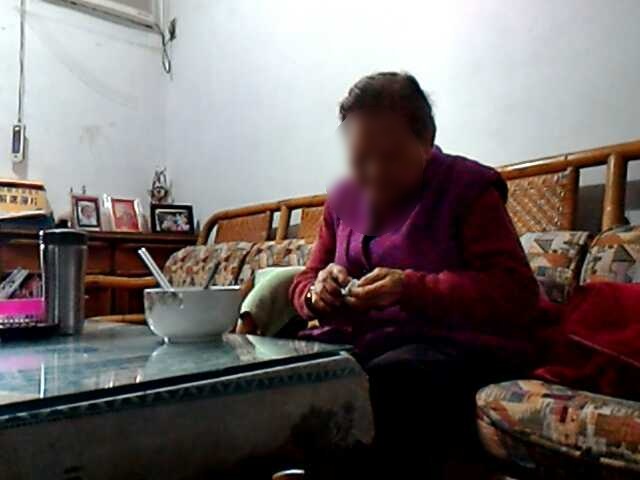} &
		\includegraphics[width=0.14\columnwidth]{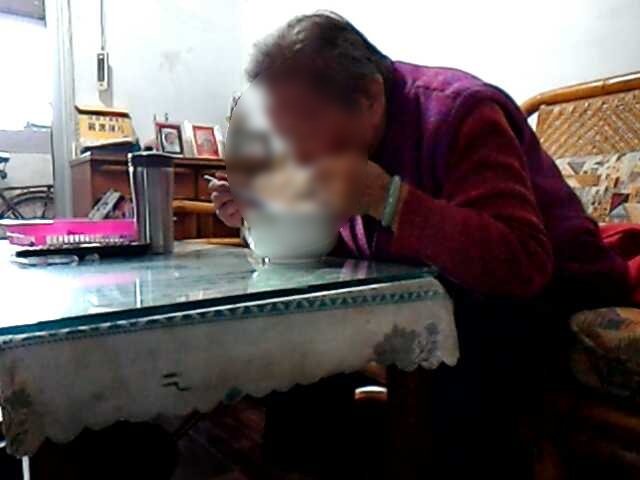} &
		\includegraphics[width=0.14\columnwidth]{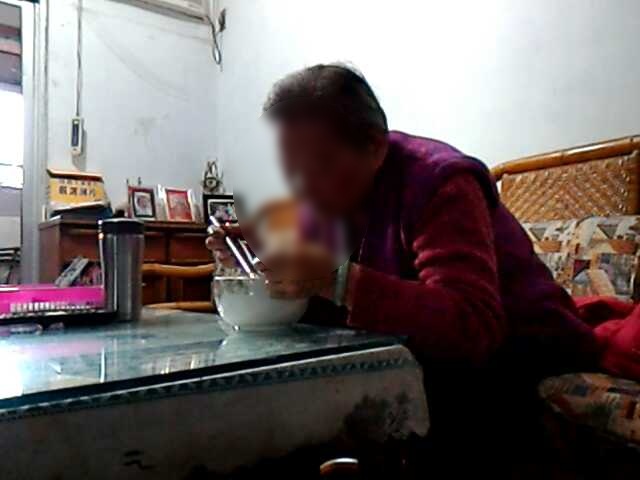} &
		\includegraphics[width=0.14\columnwidth]{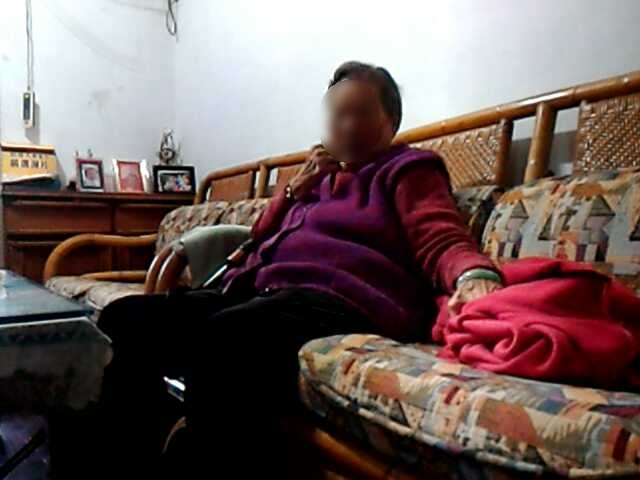} &
		\includegraphics[width=0.14\columnwidth]{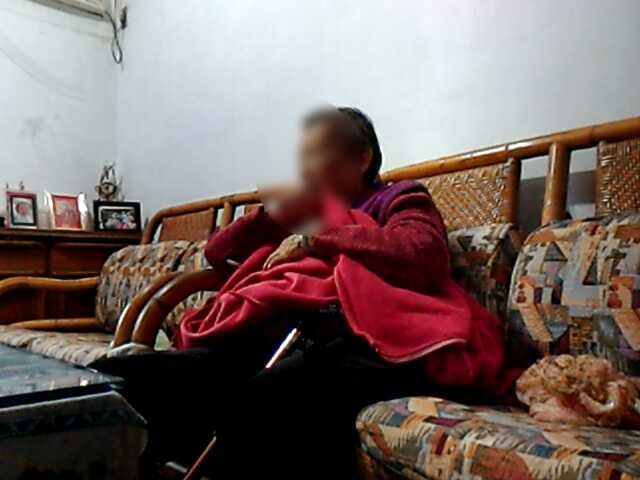} &
		\figureraisebox{VSUMM}\\
		&
		\figureraisebox{6} &
		\includegraphics[width=0.14\columnwidth]{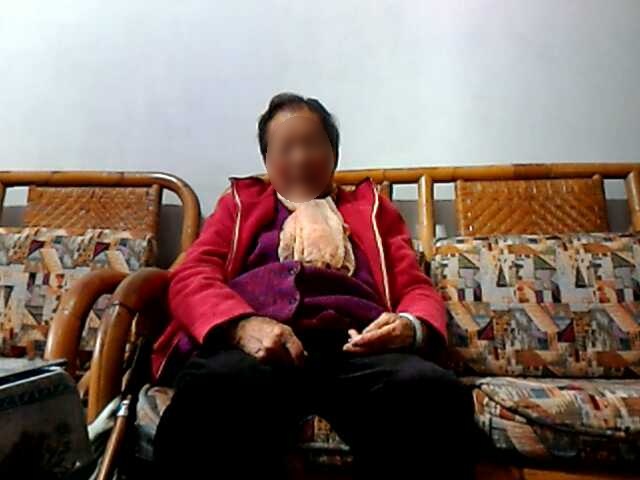} &
		\includegraphics[width=0.14\columnwidth]{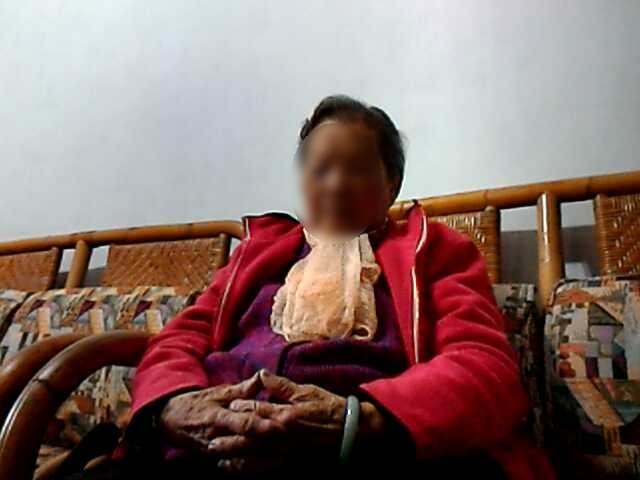} &
		\includegraphics[width=0.14\columnwidth]{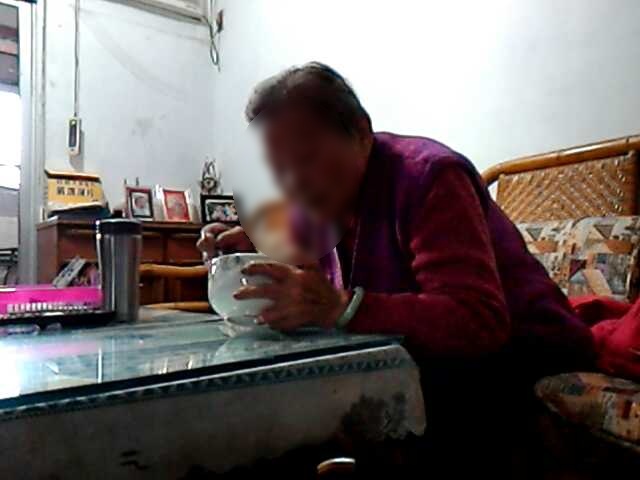} &
		\includegraphics[width=0.14\columnwidth]{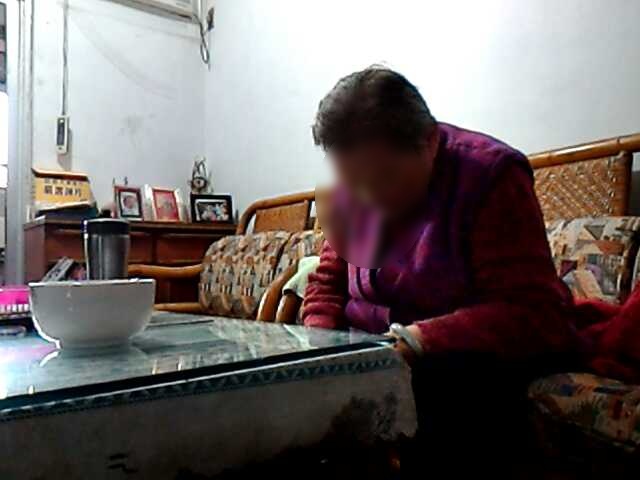} &
		\includegraphics[width=0.14\columnwidth]{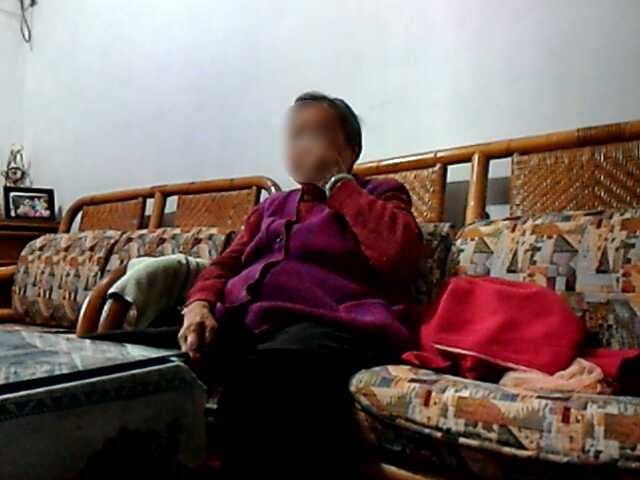} &
		\includegraphics[width=0.14\columnwidth]{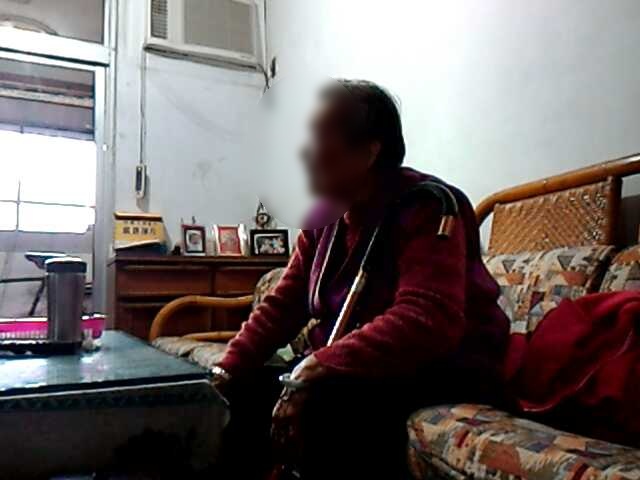} &
		\includegraphics[width=0.14\columnwidth]{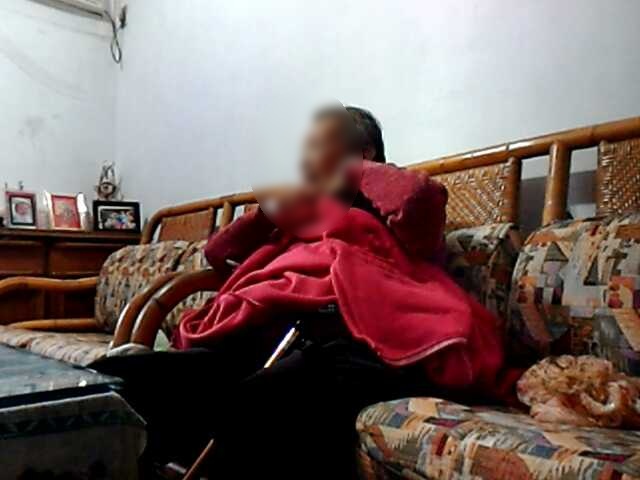} &
		\includegraphics[width=0.14\columnwidth]{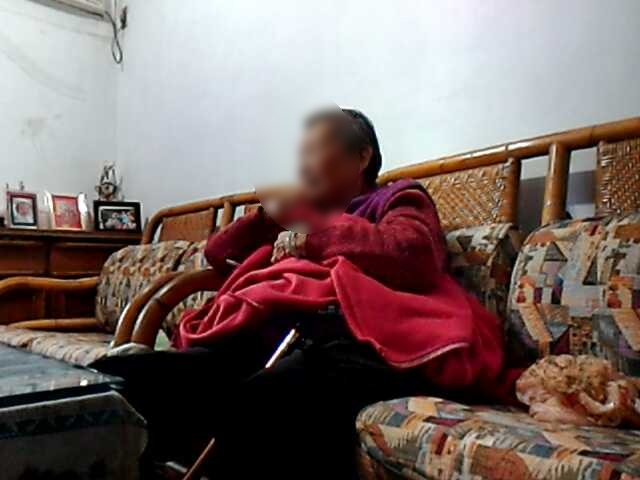} &
		\figureraisebox{DPP}\\
		&
		\figureraisebox{7} &
		\includegraphics[width=0.14\columnwidth]{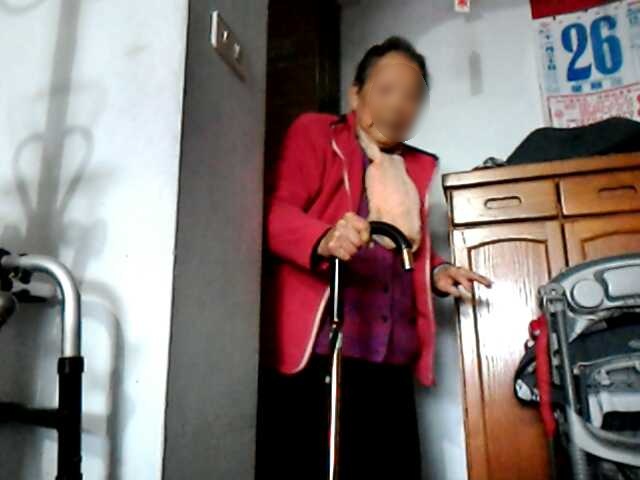} &
		\includegraphics[width=0.14\columnwidth]{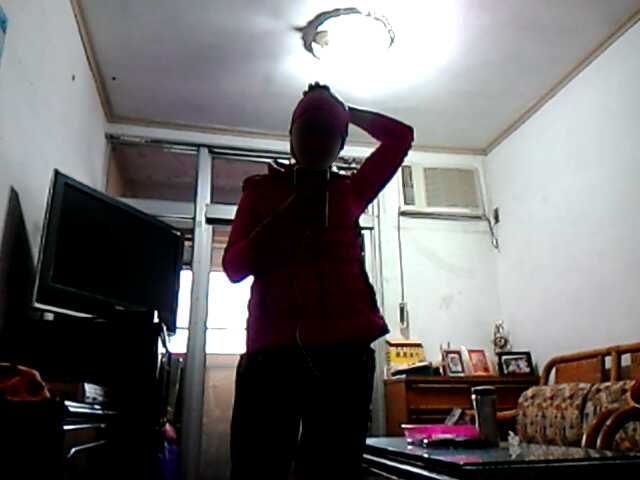} &
        \includegraphics[width=0.14\columnwidth]{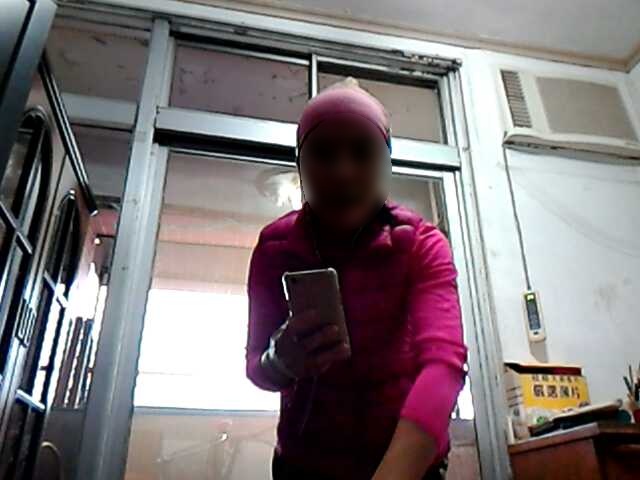} &
		\includegraphics[width=0.14\columnwidth]{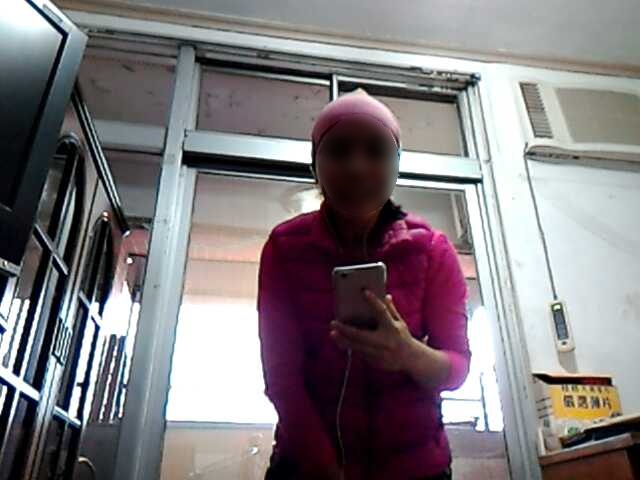} &        
		\includegraphics[width=0.14\columnwidth]{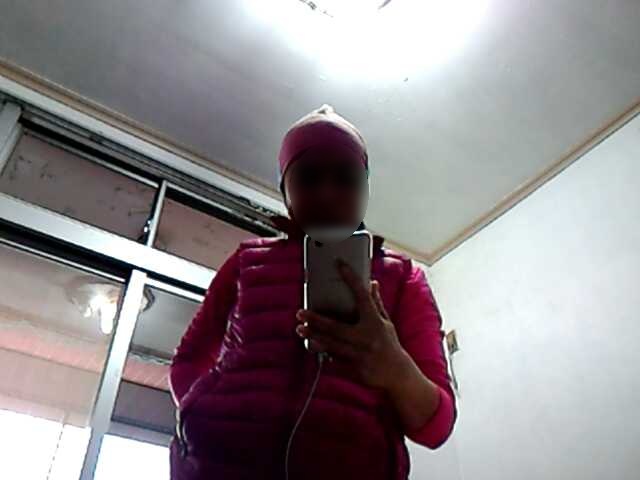} &
		\includegraphics[width=0.14\columnwidth]{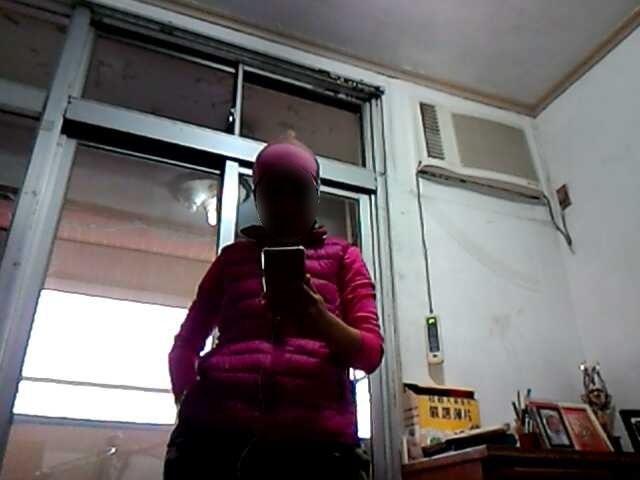} &
		\includegraphics[width=0.14\columnwidth]{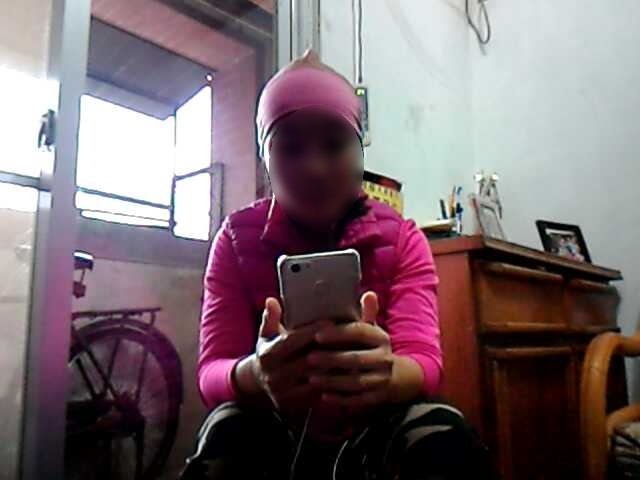} &
		\includegraphics[width=0.14\columnwidth]{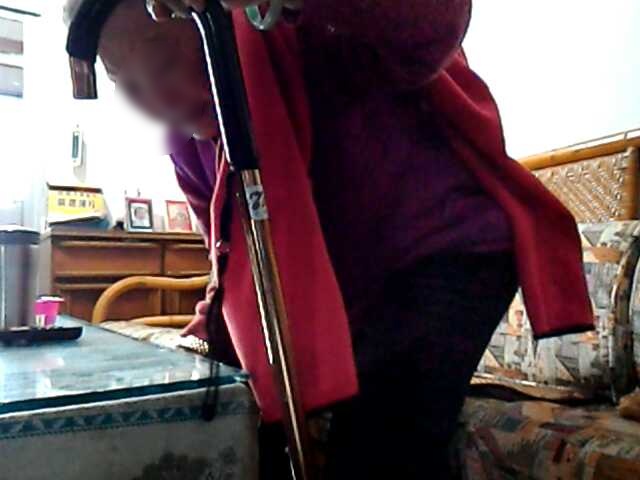} &
		\figureraisebox{DR-DSN}\\
		&
		\figureraisebox{8} &
		\includegraphics[width=0.14\columnwidth]{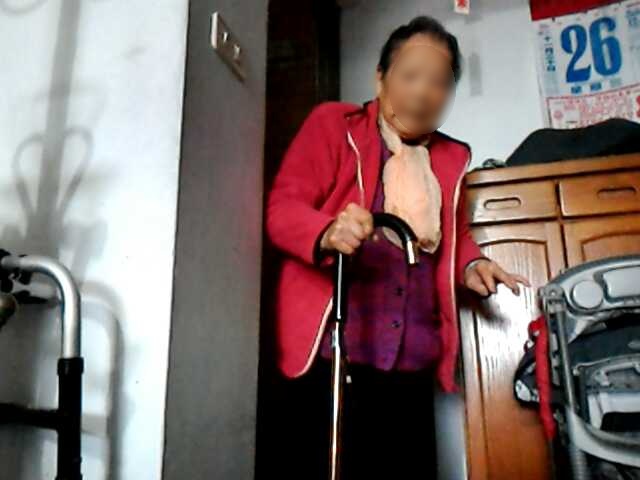} &
		\includegraphics[width=0.14\columnwidth]{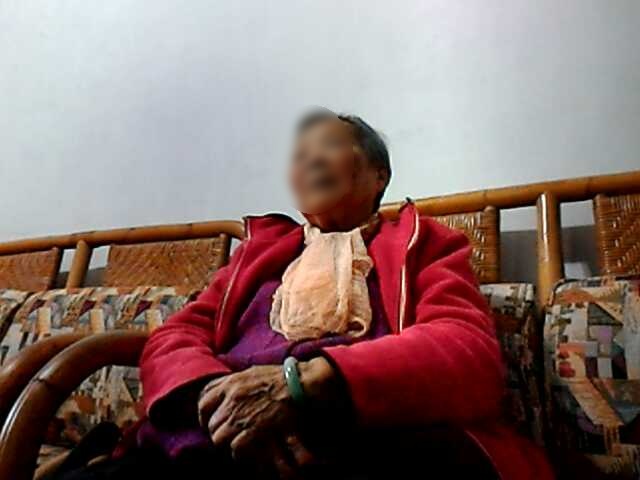} &
		\includegraphics[width=0.14\columnwidth]{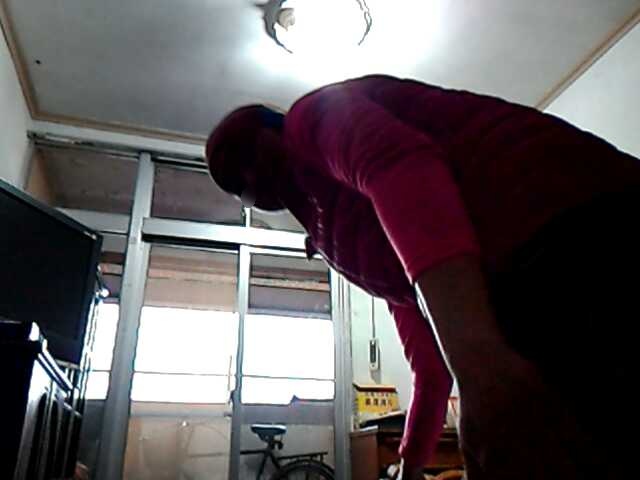} &
		\includegraphics[width=0.14\columnwidth]{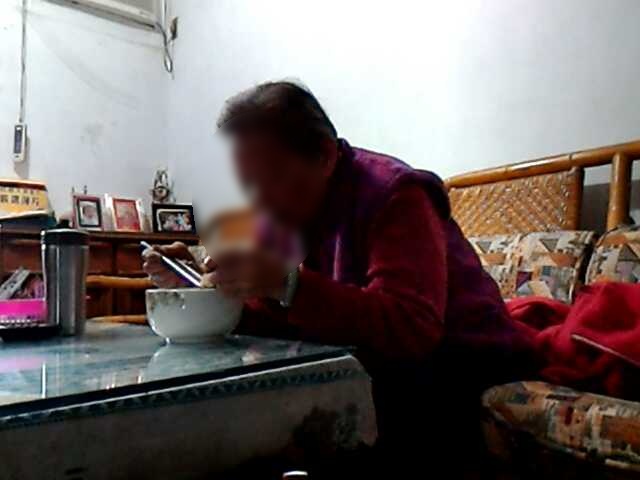} &
		\includegraphics[width=0.14\columnwidth]{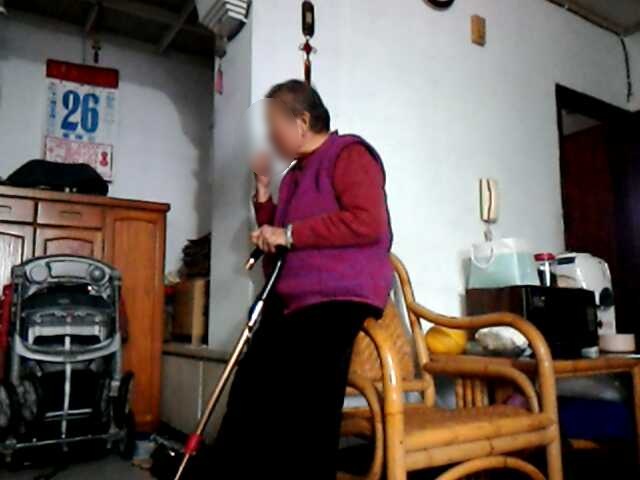} &
		\includegraphics[width=0.14\columnwidth]{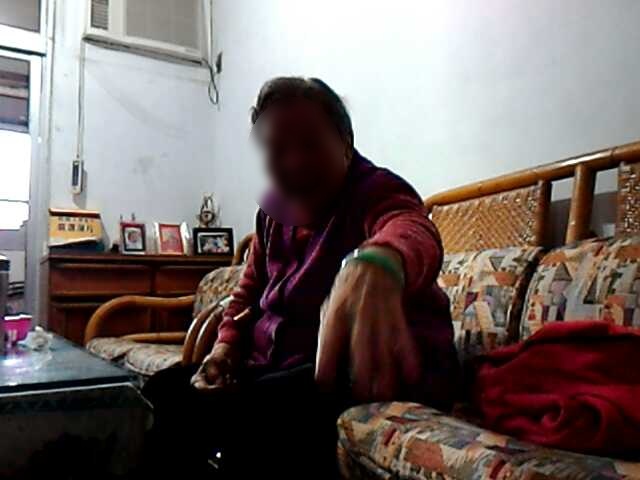} &
		\includegraphics[width=0.14\columnwidth]{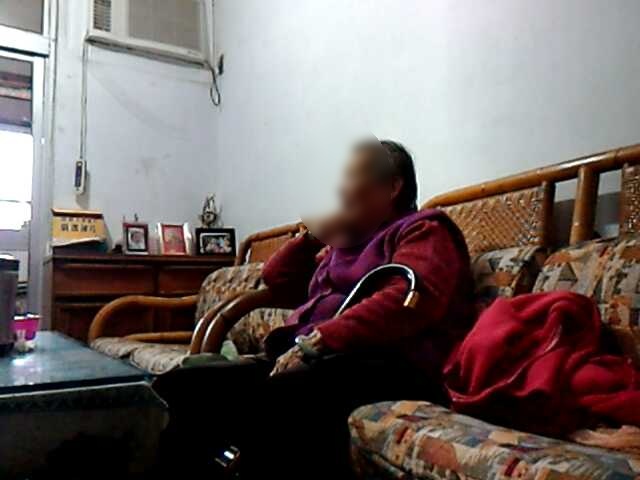} &
		\includegraphics[width=0.14\columnwidth]{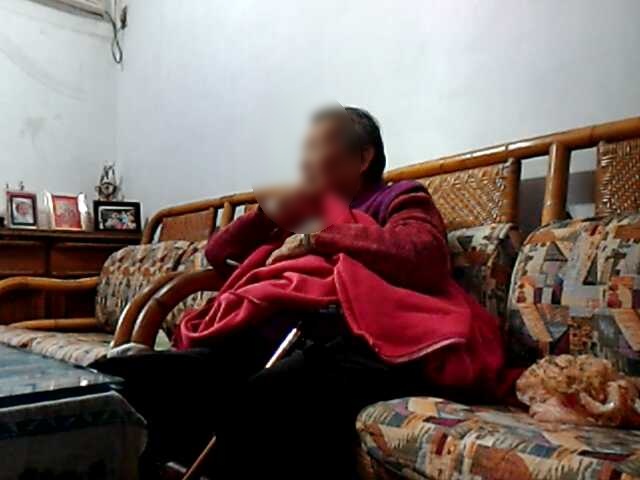} &
 		\figureraisebox{Proposed}\\		
		\rule{0pt}{0.8cm}\multirow{5}{*}[0.05cm]{\includegraphics[width=0.6\columnwidth]{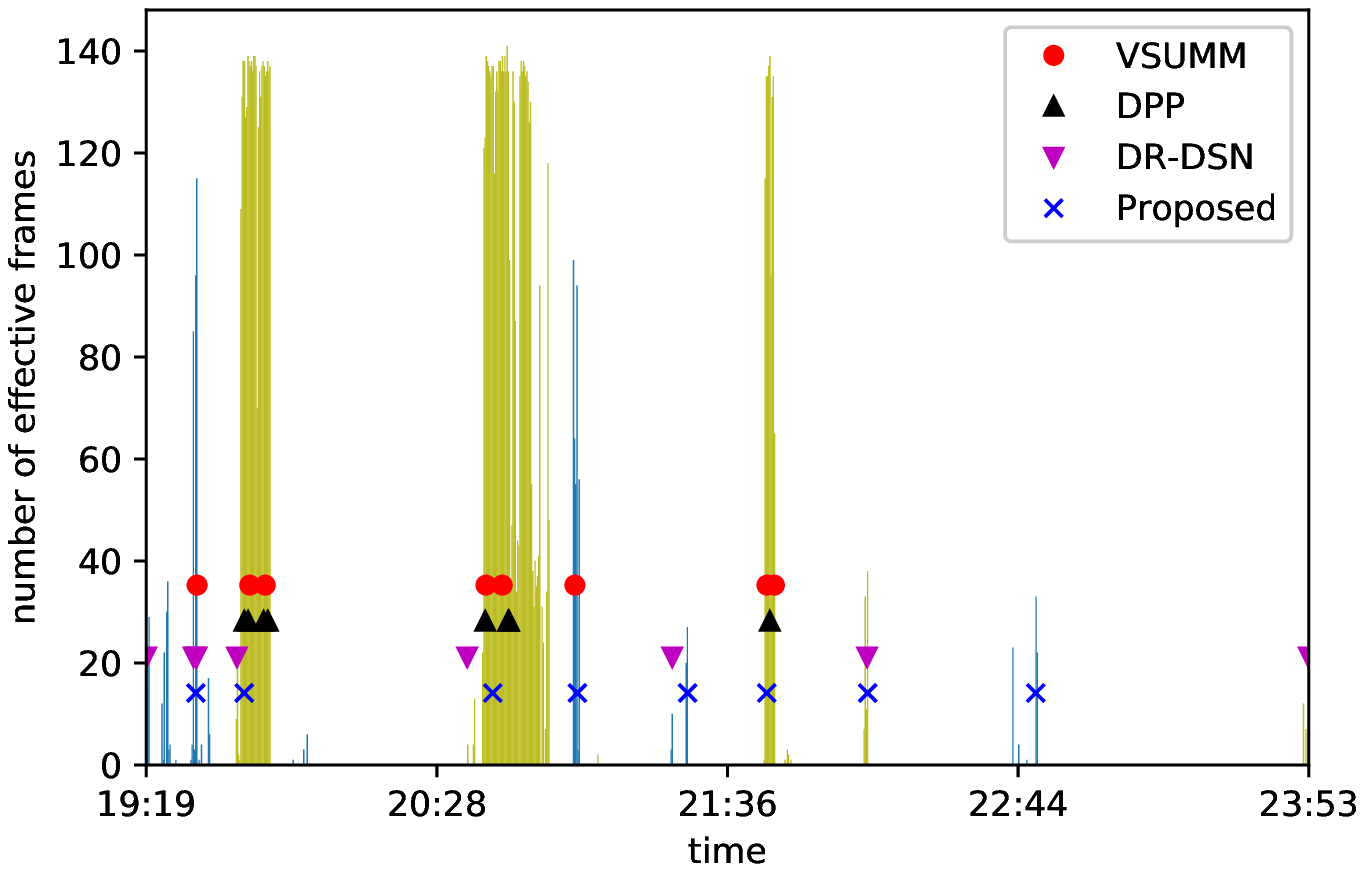}} &
		\figureraisebox{9} &
		\includegraphics[width=0.14\columnwidth]{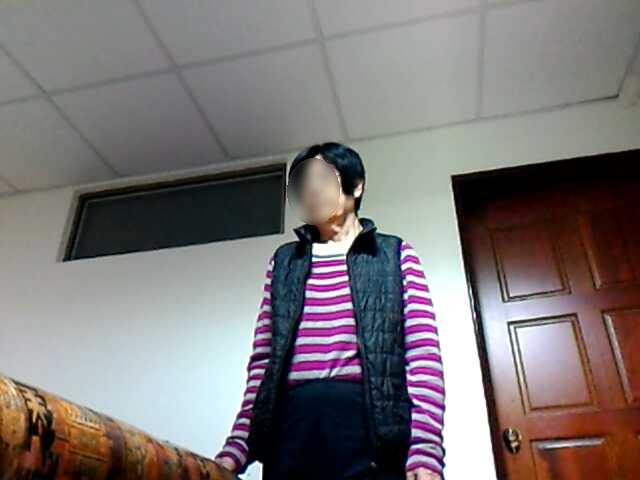} &
		\includegraphics[width=0.14\columnwidth]{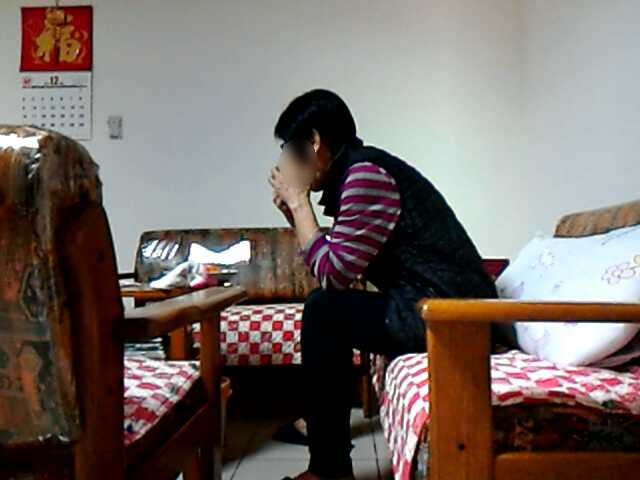} &
		\includegraphics[width=0.14\columnwidth]{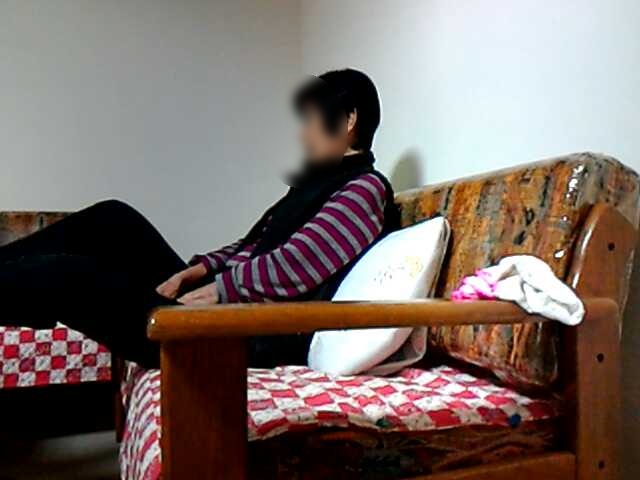} &
		\includegraphics[width=0.14\columnwidth]{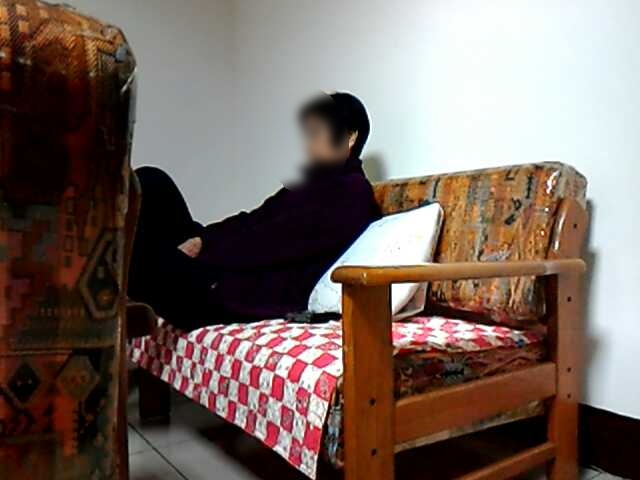} &
		\includegraphics[width=0.14\columnwidth]{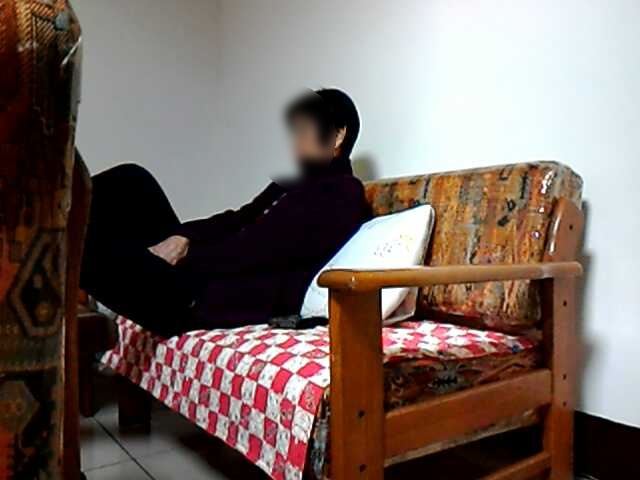} &
		\includegraphics[width=0.14\columnwidth]{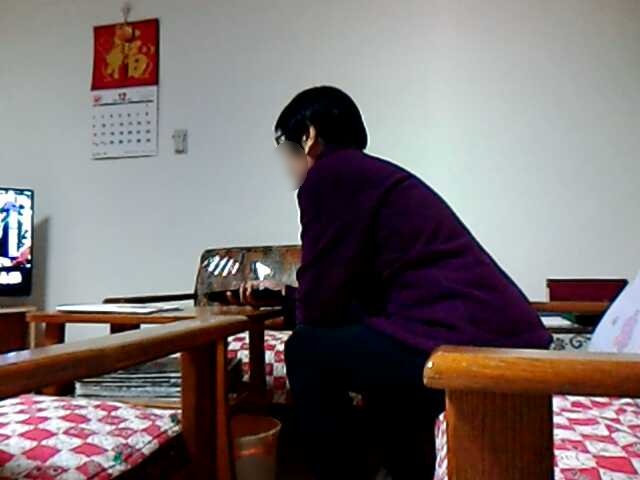} &
		\includegraphics[width=0.14\columnwidth]{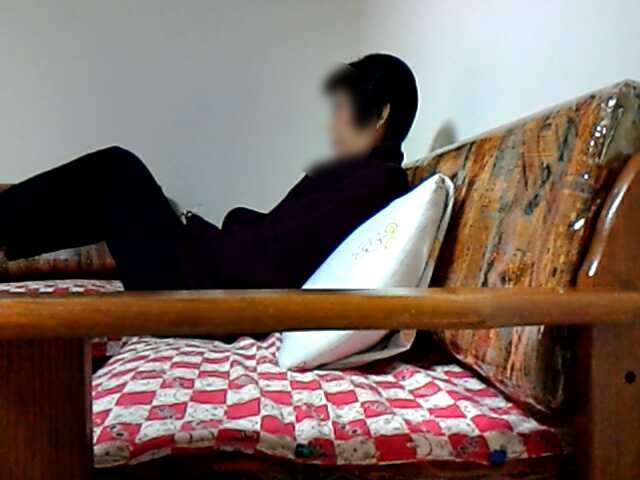} &
		\includegraphics[width=0.14\columnwidth]{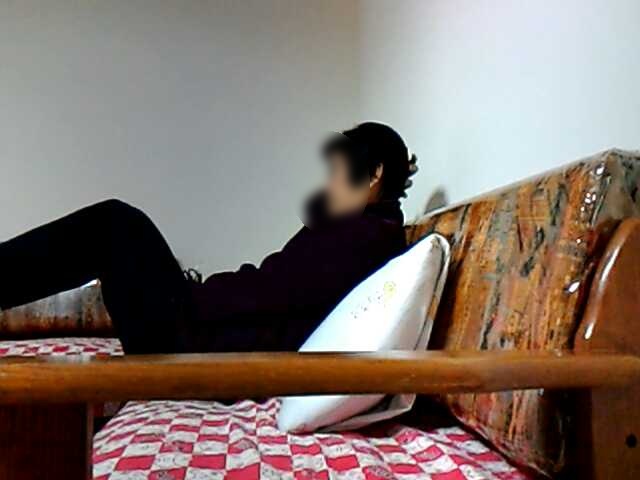} &
		\figureraisebox{VSUMM}\\
		&
		\figureraisebox{10} &
		\includegraphics[width=0.14\columnwidth]{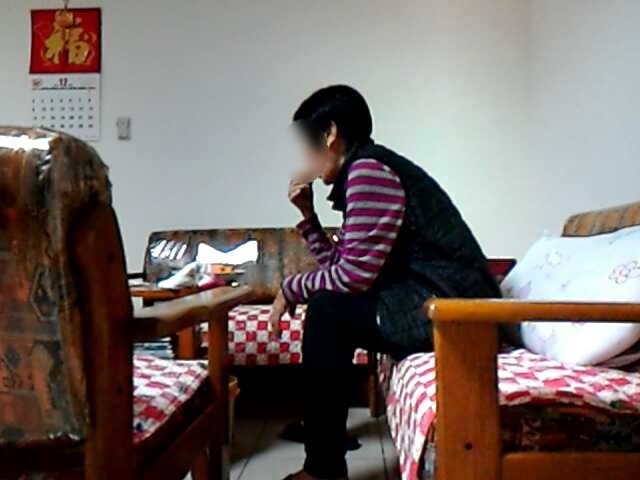} &
		\includegraphics[width=0.14\columnwidth]{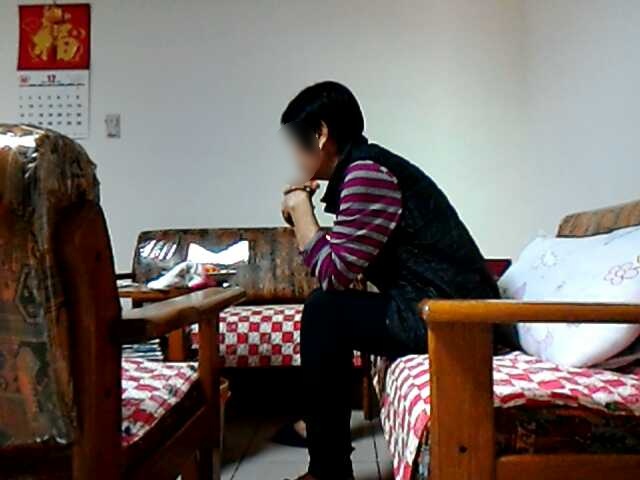} &
		\includegraphics[width=0.14\columnwidth]{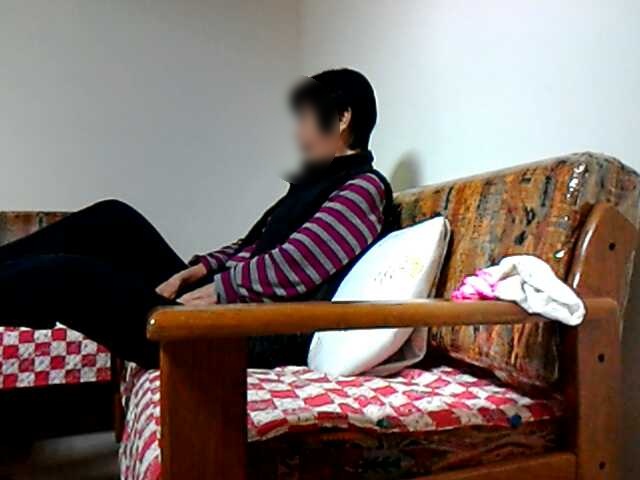} &
		\includegraphics[width=0.14\columnwidth]{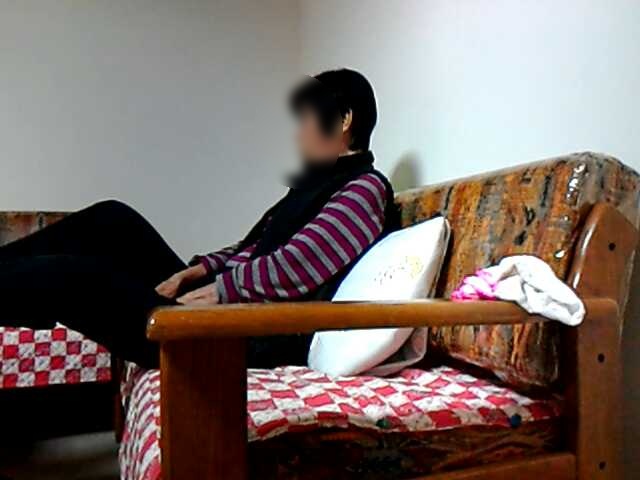} &
		\includegraphics[width=0.14\columnwidth]{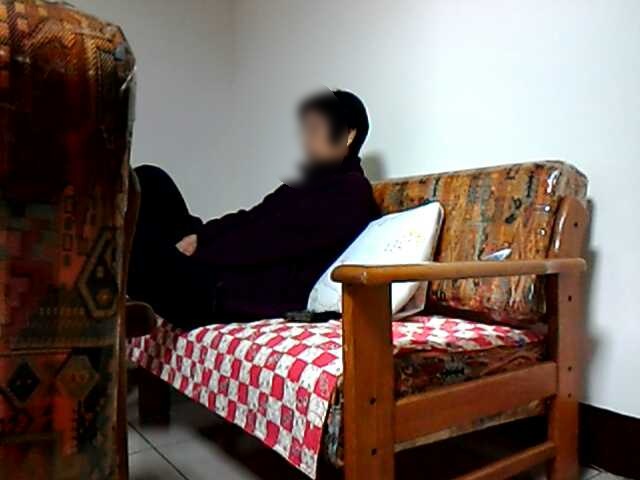} &
		\includegraphics[width=0.14\columnwidth]{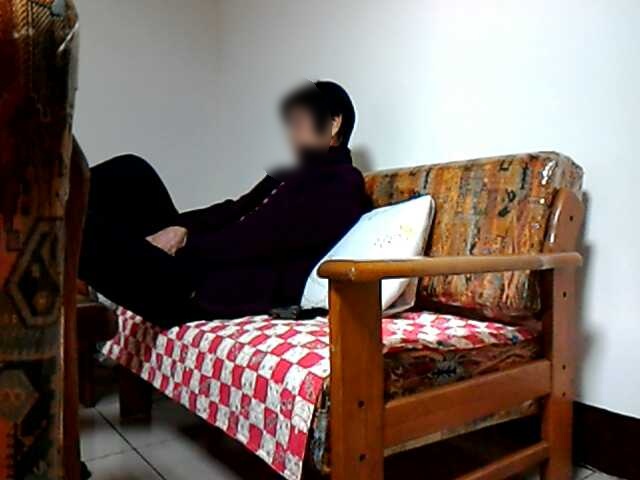} &
		\includegraphics[width=0.14\columnwidth]{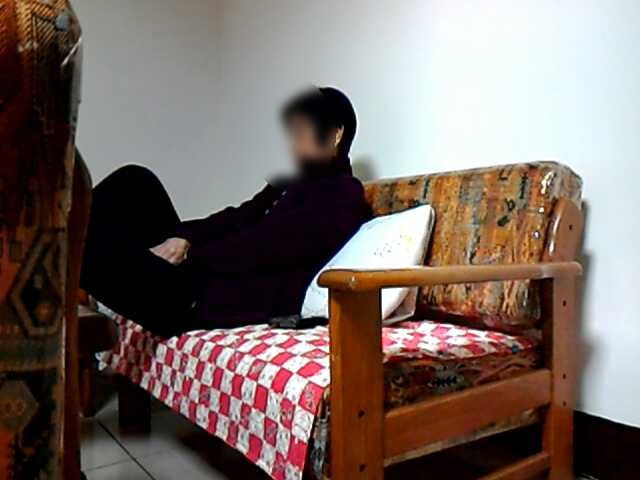} &
		\includegraphics[width=0.14\columnwidth]{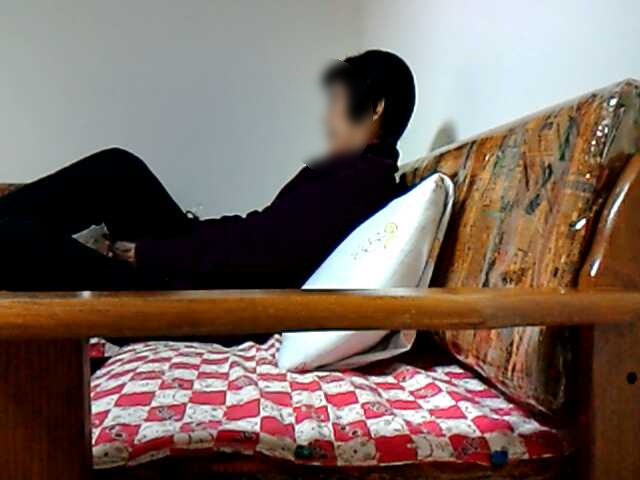} &
		\figureraisebox{DPP}\\
		&
		\figureraisebox{11} &
		\includegraphics[width=0.14\columnwidth]{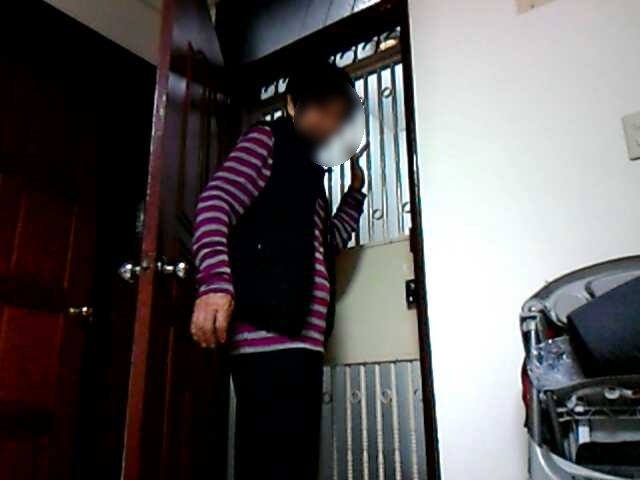} &
		\includegraphics[width=0.14\columnwidth]{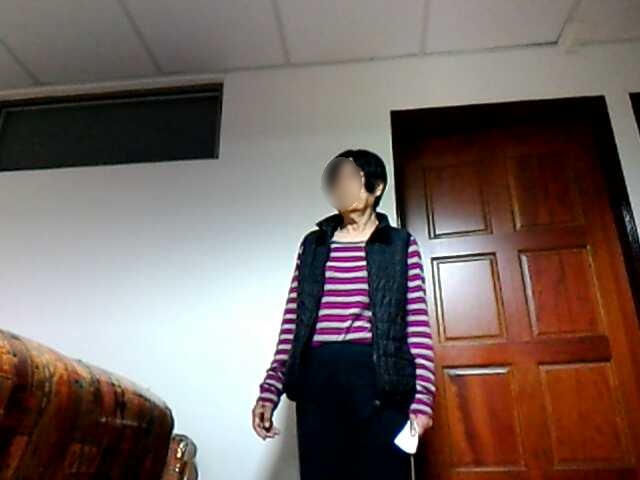} &
		\includegraphics[width=0.14\columnwidth]{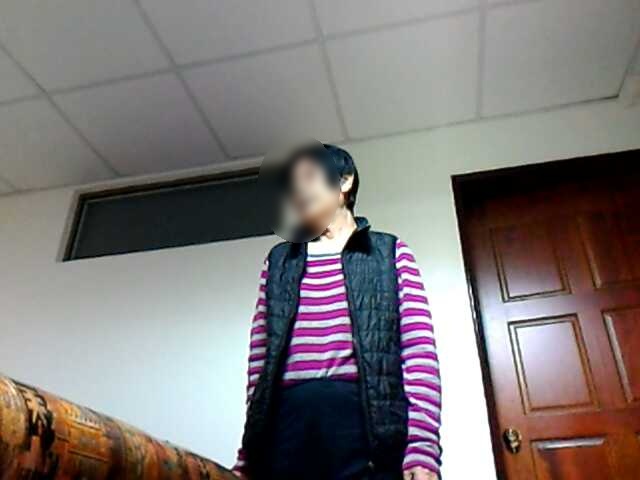} &
		\includegraphics[width=0.14\columnwidth]{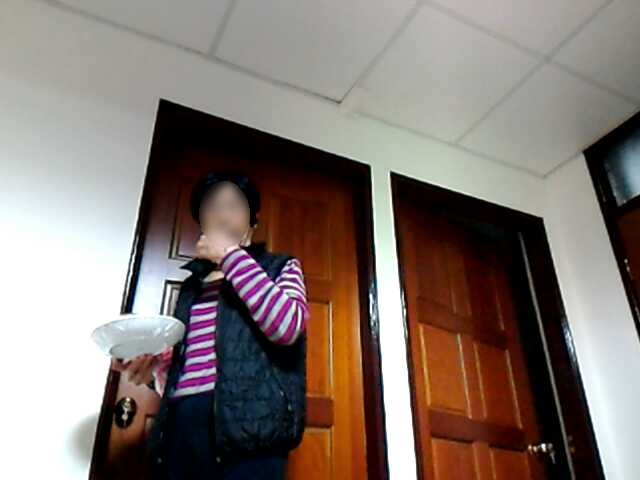} &
		\includegraphics[width=0.14\columnwidth]{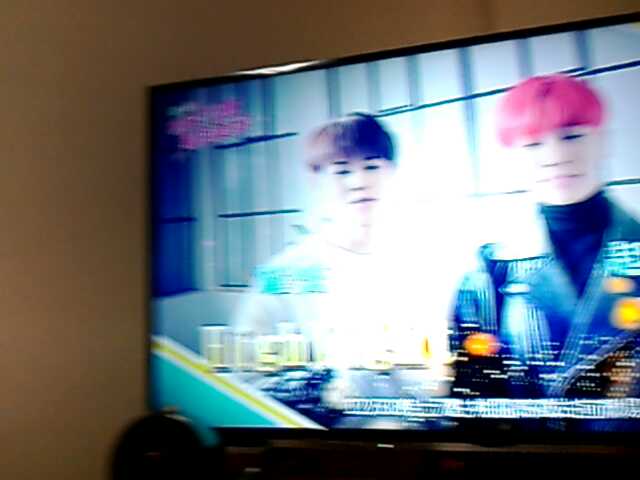} &
		\includegraphics[width=0.14\columnwidth]{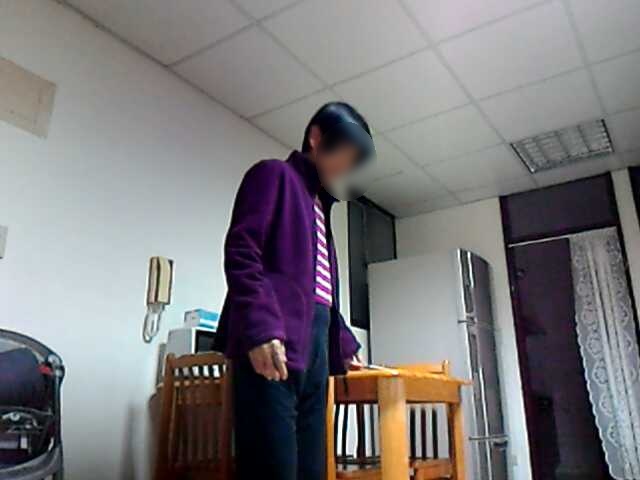} &
		\includegraphics[width=0.14\columnwidth]{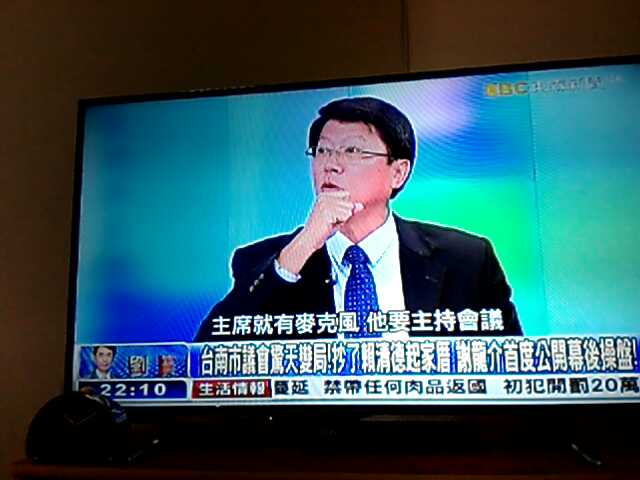} &
		\includegraphics[width=0.14\columnwidth]{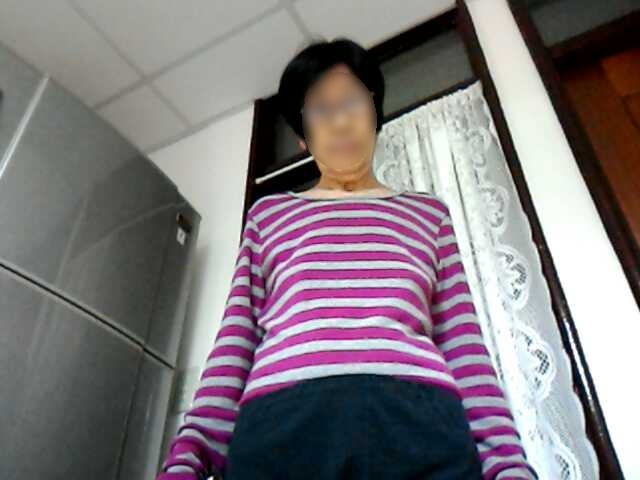} &
		\figureraisebox{DR-DSN}\\
		&
		\figureraisebox{12} &
		\includegraphics[width=0.14\columnwidth]{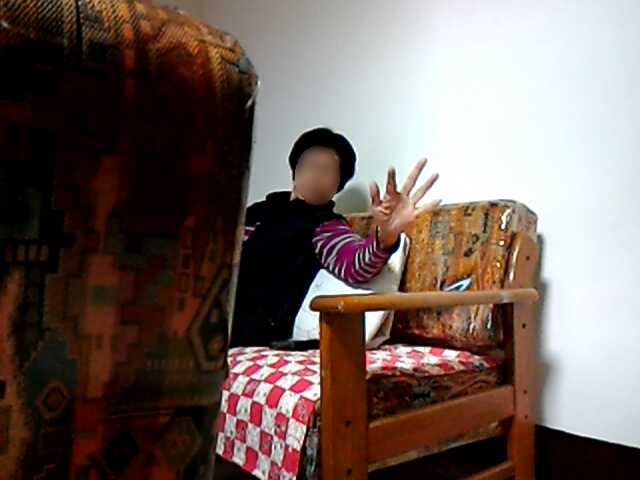} &
		\includegraphics[width=0.14\columnwidth]{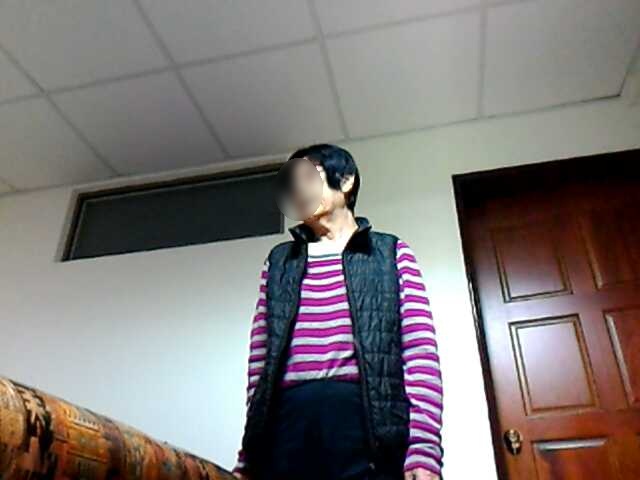} &
		\includegraphics[width=0.14\columnwidth]{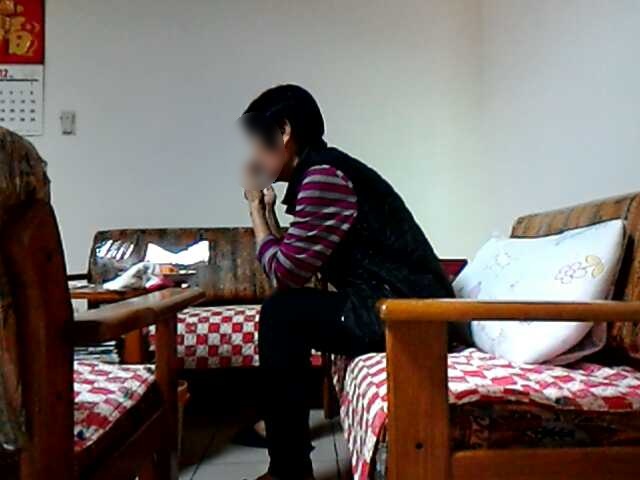} &
		\includegraphics[width=0.14\columnwidth]{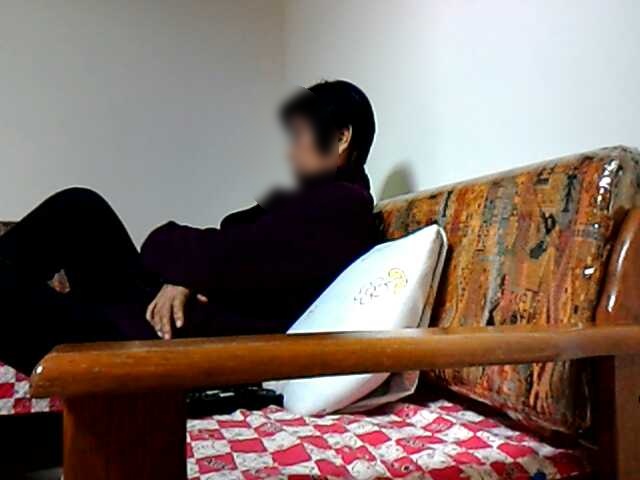} &
		\includegraphics[width=0.14\columnwidth]{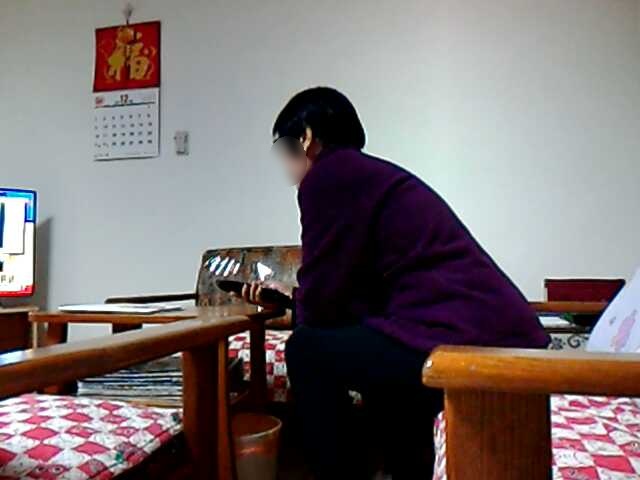} &
		\includegraphics[width=0.14\columnwidth]{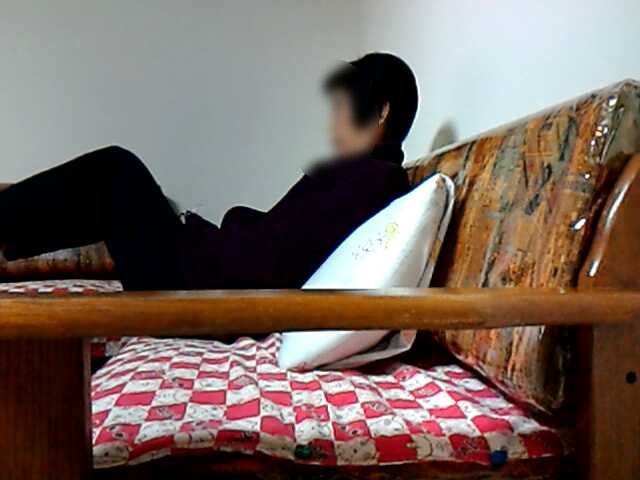} &
		\includegraphics[width=0.14\columnwidth]{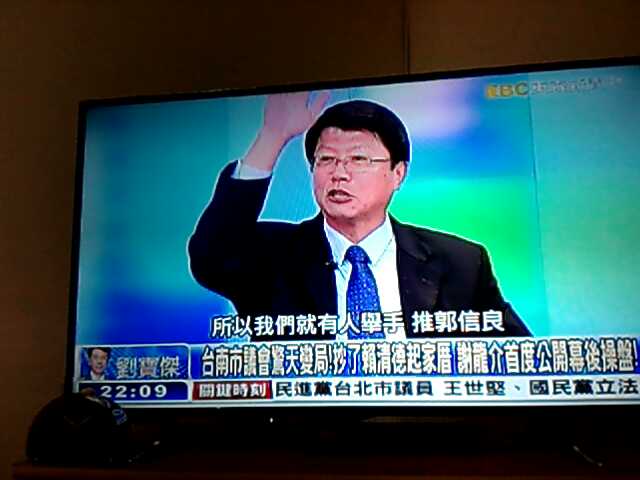} &
		\includegraphics[width=0.14\columnwidth]{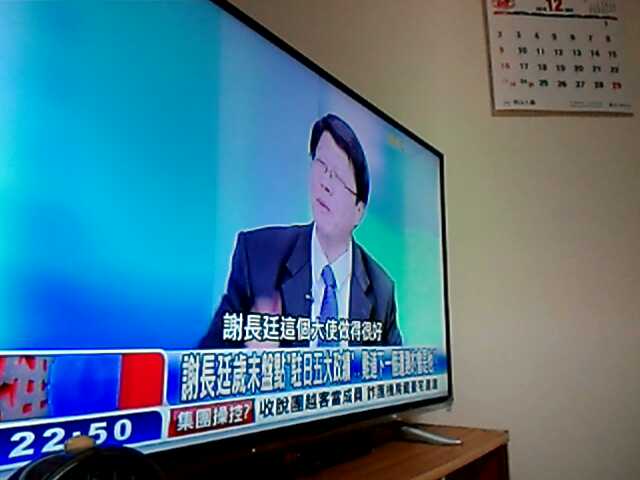} &
		\figureraisebox{Proposed}\\
		& & (a) & (b) & (c) & (d) & (e) & (f) & (g) & (h) & Method\\
	\end{tabular}
	\caption{Distribution charts and keyframes.  From top to bottom: Videos 1, 2, and 3. At left: Histograms of well-posed frames of the three videos in 1000 bins. We draw the bars in two alternative colors to show the clusters generated by the proposed method. The numbers of clusters of the three videos are 11, 8, and 12. Note \hl{that} the compared methods are not affected by the clusters and some clusters are invisible because they contains too few frames to be drawn under the chart resolution. The marks indicate the time of the keyframes and we use evenly separated heights for the ease of observation. At right: selected keyframes arranged in temporal order. The proposed method does not generate repeating keyframes such as 1(ab), 1(fg), 2(efgh), 5(cd), 6(ef), 7(de), 7(gh), 10(de), 11(cd), and 11(efg). \hl{We blur participants' faces to protect their privacy.}}
	\label{fig:Experiment}
\end{figure*}

Qualitative comparisons are enclosed in Figure~\ref{fig:Experiment}. A set of 8 key frames (a-h) selected by the 4 different methods are shown on the right while the histogram of well-posed frames and the time of keyframes selection are shown on the left. The proposed method prevents the problem of consecutive similar keyframes arising in other methods because it exploits timestamps and a threshold $h$ to partition frames into temporally disjoint clusters, which are separated from each other with a gap that tends to isolate different activities. In contrast, VSUMM and DPP ignore temporal information and \hl{their} keyframes are not \hl{sufficiently diverse}. DR-DSN generates summaries containing diverse keyframes but neglects representative events of eating a lunch box in video 1, and sitting on a sofa in videos 2 and 3. 
The proposed method generates diverse and representative summaries, but we have not fully implemented the person identification component, which results in a person on TV shown in video 3's summary, as the images 12(g)(h) of Figure~\ref{fig:Experiment}. \hl{Our code is available at} \url{https://github.com/yangchihyuan/RobotVideoSummary}.
\section{Conclusion and Future Study}
The paper presents an effective method to generate video summaries using a social robot for family members to care about seniors living alone. We use a pose estimation method to detect humans to control the robot's movements to capture well-posed frames. We use human pose and image quality information to disregard ill-posed frames and develop a summarization method to generate diverse and representative summaries. Experimental results show that our summaries prevent the problems of \hl{redundancy} and unrepresentative keyframes generated by existing methods.

Our immediate plan is to integrate a person identification algorithm into our system to ensure the robot to keep \hl{track} on a target user. During our experiment, our users express strong demand for fall detection and immediate notification, which are important features. In addition, we plan to take preferences into \hl{account} to create personalized summaries.
\section{Acknowledgements}
\hl{This research was supported in part by the Ministry of Science and Technology of Taiwan (MOST 108-2633-E-002-001, 107-2218-E-002-009, 107-2811-E-002-575), National Taiwan University (NTU-108L104039), Intel Corporation, Delta Electronics and Compal Electronics.}
\balance{} 

\bibliographystyle{SIGCHI-Reference-Format}
\bibliography{MobileHCI}


\begin{thebibliography}{00}


\ifx \showCODEN    \undefined \def \showCODEN     #1{\unskip}     \fi
\ifx \showDOI      \undefined \def \showDOI       #1{{\tt DOI:}\penalty0{#1}\ }
  \fi
\ifx \showISBNx    \undefined \def \showISBNx     #1{\unskip}     \fi
\ifx \showISBNxiii \undefined \def \showISBNxiii  #1{\unskip}     \fi
\ifx \showISSN     \undefined \def \showISSN      #1{\unskip}     \fi
\ifx \showLCCN     \undefined \def \showLCCN      #1{\unskip}     \fi
\ifx \shownote     \undefined \def \shownote      #1{#1}          \fi
\ifx \showarticletitle \undefined \def \showarticletitle #1{#1}   \fi
\ifx \showURL      \undefined \def \showURL       #1{#1}          \fi

\bibitem{Cao17_CVPR_OpenPose}
{Zhe Cao}, {Tomas Simon}, {Shih-En Wei}, {and} {Yaser Sheikh}. 2017.
\newblock \showarticletitle{Realtime Multi-Person {2D} Pose Estimation using
  Part Affinity Fields}. In {\em CVPR}.
\newblock


\bibitem{Chen18_FG}
{Chaona Chen}, {Oliver G.~B. Garrod}, {Jiayu Zhan}, {Jonas Beskow Philippe~G.
  Schyns}, {and} {Rachael~E. Jack}. 2018.
\newblock \showarticletitle{Reverse Engineering Psychologically Valid Facial
  Expressions of Emotion into Social Robots}. In {\em FG}.
\newblock


\bibitem{Avila11}
{Sandra Eliza~Fontes de Avila}, {Ana Paula~Brand\ {a}o Lopes}, {Antonio da
  Luz~Jr.}, {and} {Arnaldo de Albuquerque~Ara\'{u}jo}. 2011.
\newblock \showarticletitle{{VSUMM}: A mechanism designed to produce static
  video summaries and a novel evaluation method}.
\newblock {\em Pattern Recognition Letters\/} {32}, 1 (2011), 56--68.
\newblock


\bibitem{Gong14}
{Boqing Gong}, {Wei-Lun Chao}, {Kristen Grauman}, {and} {Fei Sha}. 2014.
\newblock \showarticletitle{Diverse Sequential Subset Selection for Supervised
  Video Summarization}. In {\em NIPS}.
\newblock


\bibitem{Iandola16_SqueezeNet}
{Forrest~N. Iandola}, {Matthew~W. Moskewicz}, {Khalid Ashraf}, {Song Han},
  {William~J. Dally}, {and} {Kurt Keutzer}. 2016.
\newblock \showarticletitle{SqueezeNet: AlexNet-level accuracy with 50x fewer
  parameters and {\textless}1MB model size}.
\newblock {\em arXiv:1602.07360\/} (2016).
\newblock


\bibitem{OpenVINO}
{Intel}. 2019.
\newblock Distribution of {OpenVINO} Toolkit.
\newblock \url{https://software.intel.com/en-us/openvino-toolkit}.   (2019).
\newblock


\bibitem{Klamer10}
{Tineke Klamer} {and} {Soumaya {Ben Allouch}}. 2010.
\newblock \showarticletitle{Acceptance and use of a social robot by elderly
  users in a domestic environment}. In {\em Proceedings of IEEE International
  Conference on Pervasive Computing Technologies for Healthcare}.
\newblock


\bibitem{PechPacheco2000_DiatomAI}
{Jos{\'e}~Luis Pech-Pacheco}, {Gabriel Crist{\'o}bal}, {Jes{\'u}s
  Chamorro-Mart{\'i}nez}, {and} {Joaqu{\'i}n Fern{\'a}ndez-Valdivia}. 2000.
\newblock \showarticletitle{Diatom Autofocusing in Brightfield Microscopy: a
  Comparative Study}. In {\em ICPR}.
\newblock


\bibitem{Safriaz18}
{M.~Saquib Sarfraz}, {Arne Schumann}, {Andreas Eberle}, {and} {Rainer
  Stiefelhagen}. 2018.
\newblock \showarticletitle{A Pose-Sensitive Embedding for Person
  Re-Identification with Expanded Cross Neighborhood Re-Ranking}. In {\em
  CVPR}.
\newblock


\bibitem{Sigurdsson16}
{Gunnar~A. Sigurdsson}, {G\"{o}l Varol}, {Xiaolong Wang}, {Ali Farhadi}, {Ivan
  Laptev}, {and} {Abhinav Gupta}. 2016.
\newblock \showarticletitle{Hollywood in Homes: Crowdsourcing Data Collection
  for Activity Understanding}. In {\em ECCV}.
\newblock


\bibitem{Tan18_IJSR}
{Zheng-Hua Tan}, {Nicolai~B{\ae}k Thomsen}, {Xiaodong Duan}, {Evgenios
  Vlachos}, {Sven~Ewan Shepstone}, {Morten~H{\o}jfeldt Rasmussen}, {and}
  {Jesper~Lisby H{\o}jvang}. 2018.
\newblock \showarticletitle{{iSocioBot}: A Multimodal Interactive Social
  Robot}.
\newblock {\em IJSR\/} {10}, 1 (2018), 5--19.
\newblock


\bibitem{Zhang16_ECCV}
{Ke Zhang}, {Wei-Lun Chao}, {Fei Sha}, {and} {Kristen Grauman}. 2016.
\newblock \showarticletitle{Video summarization with long short-term memory}.
  In {\em ECCV}.
\newblock


\bibitem{Zhou18_AAAI}
{Kaiyang Zhou}, {Yu Qiao}, {and} {Tao Xiang}. 2018.
\newblock \showarticletitle{Deep Reinforcement Learning for Unsupervised Video
  Summarization with Diversity-Representativeness Reward}. In {\em AAAI}.
\newblock


\end{thebibliography}

\end{document}